\documentclass{article}

\PassOptionsToPackage{numbers, compress}{natbib}
\usepackage[preprint]{neurips_2025}

\usepackage[utf8]{inputenc} % allow utf-8 input
\usepackage[T1]{fontenc}    % use 8-bit T1 fonts
\usepackage{hyperref}       % hyperlinks
\usepackage{url}            % simple URL typesetting
\usepackage{booktabs}       % professional-quality tables
\usepackage{amsfonts}       % blackboard math symbols
\usepackage{nicefrac}       % compact symbols for 1/2, etc.
\usepackage{microtype}      % microtypography        % colors
\usepackage{amsmath}
\usepackage{enumitem}
\usepackage{graphicx}
\usepackage{subcaption}
\usepackage{caption} 
\usepackage{float}
\usepackage{geometry}

\usepackage[table]{xcolor}
\usepackage{amssymb}
\usepackage{booktabs}
\usepackage{pifont}
\usepackage{colortbl}
\usepackage{multirow}

\title{Beyond Editing Pairs: Fine-Grained Instructional Image Editing via Multi-Scale Learnable Regions}

\author{%
  Chenrui Ma \\
  Department of Electronic Engineering and Computer Science\\
  University of California, Irvine\\
  Irvine, CA 92612 \\
  \texttt{chenrum@uci.edu} \\
  \And
  Xi Xiao \\
  Department of Computer Science\\
  University of Alabama at Birmingham\\
  Birmingham, AL 35294 \\
  \texttt{xxiao@uab.edu} \\
  \AND
  Tianyang Wang \\
  Department of Computer Science\\
  University of Alabama at Birmingham\\
  Birmingham, AL 35294 \\
  \texttt{tw2@uab.edu} \\
  \AND
  Yanning Shen \\
  Department of Electronic Engineering and Computer Science\\
  University of California, Irvine\\
  Irvine, CA 92612 \\
  \texttt{yannings@uci.edu} \\
}

\begin{document}

\maketitle

\begin{abstract}

Current text-driven image editing methods typically follow one of two directions: relying on large-scale, high-quality editing pair datasets to improve editing precision and diversity, or exploring alternative dataset-free techniques. However, constructing large-scale editing datasets requires carefully designed pipelines, is time-consuming, and often results in unrealistic samples or unwanted artifacts. Meanwhile, dataset-free methods may suffer from limited instruction comprehension and restricted editing capabilities.
Faced with these challenges, the present work develops a novel paradigm for instruction-driven image editing that leverages widely available and enormous text-image pairs, instead of relying on editing pair datasets. Our approach introduces a multi-scale learnable region to localize and guide the editing process. By treating the alignment between images and their textual descriptions as supervision and learning to generate task-specific editing regions, our method achieves high-fidelity, precise, and instruction-consistent image editing.
Extensive experiments demonstrate that the proposed approach attains state-of-the-art performance across various tasks and benchmarks, while exhibiting strong adaptability to various types of generative models.
% Code will be release after publish

% Current text-driven image editing methods are either pursuing large-scale and high-quality editing pairs datasets to achieve more precise and diverse editing performance, or exploring training-free editing techniques. However, construct large-scale image editing dataset needs careful pipeline design and usually suffers from unrealistic data samples and unwanted artifacts and time consuming. While training-free editing techniques like inversion are limited in instruction comprehension and quantity of editing operations. In this paper, we propose a new paradigm that enable instruction-driven image editing using widely-consistence and accessible text-image pairs data instead of editing pairs data via multi-scale learnable region. By serving the alignment between image and text description as editing supervision and producing a learnable editing region, we achieve high-fidelity, precise and instruction-followed image editing. Various experiments demonstrate that our method achieve comparable SOTA performance across tasks and benchmarks, and additionally highlight the comparability of our method with various generative models.
\end{abstract}
    
\section{Introduction} \label{intro}

% With the rapid emergence of massive text-image paired data driven by social media development and electronic data recording, deep generative models have gained significant advantages in image generation \cite{rombach2022high, schuhmann2022laion}. These models are capable of synthesizing images realistic enough to deceive human recognition, making them indistinguishable from real photographs \cite{podell2023sdxlimprovinglatentdiffusion}. 
Image editing, as an important application of these technologies, empowers people to create and manipulate specific visual content in realistic images without requiring expert skills. Among various image editing methods, text-driven image editing stands out as the most accessible approach, as it allows users to modify images using natural language instructions in a simple and intuitive way \cite{Brooks_2023_CVPR, Guo_2024_CVPR}.

% {\color{blue}The following paragraph should be shortened and put in related work.}
Based on the textual prompts provided, text-driven image editing methods can generally be categorized into two groups: 1) \emph{description-driven} and 2) \emph{instruction-driven} editing.
For \textit{description-driven image editing}, users provide an original image along with a descriptive text, such as "a running dog" or "a tree with many flowers," and the model modifies the visual content accordingly \cite{Lin_2024_CVPR, Goel_2024_CVPR, Cao_2023_ICCV}. Due to the absence of explicitly specified editing objects in these descriptions, such methods typically result in global or large-scale image alterations, lacking direct and precise local editing capabilities \cite{Lin_2024_CVPR, couairon2022diffeditdiffusionbasedsemanticimage}. 
In contrast, \textit{instruction-driven image editing} utilizes clear instructional prompts, such as ``change the dog into a cat'' or ``add some cherries to the dish,'' directing the model explicitly to perform the desired edits \cite{Brooks_2023_CVPR, NEURIPS2023_64008fa3, Guo_2024_CVPR}. Compared to description-driven methods, instruction-driven editing enables more precise and targeted operations, allowing for concise and intuitive user interactions \cite{Li_2024_CVPR}. However, these methods heavily depend on high-quality instruction-editing datasets, which typically consist of an original image, an instruction prompt, and the corresponding edited image (some datasets also include masks indicating the edited regions) \cite{NEURIPS2023_64008fa3}. Such datasets are often constructed through carefully designed pipelines to obtain edited images that serve as ground truth \cite{Brooks_2023_CVPR}. 
Despite these efforts, models trained on such datasets often produce unrealistic or excessive edits and lack editing diversity due to dataset limitations \cite{Liu_2024_CVPR}. While manual filtering can improve data quality, it is labor-intensive and constrained by dataset size \cite{NEURIPS2024_05a30a0f, NEURIPS2023_64008fa3}. Additionally, editing in latent space may distort unedited regions and lack precision \cite{Li_2024_CVPR}. Faced with these challenges, this work introduces multi-scale learnable regions to localize edits and enable instructional, fine-grained, precise modifications.
% \shen{So what? How does this section connect with your work?}

% Beyond the differences between these two paradigms of text-driven image editing, there is an important phenomenon that calls for attention. 
Due to the nature of text-image pair data generation, collecting images along with corresponding descriptive texts is significantly easier than collecting image editing pairs \cite{CLIP}. For instance, the InstructPix2Pix dataset\cite{Brooks_2023_CVPR} contains approximately 0.45 million samples, and the recent large-scale edited dataset UltraEdit\cite{NEURIPS2024_05a30a0f} consists of around 4 million editing samples. In contrast, large-scale text-image datasets such as LAION-5B\cite{schuhmann2022laion} contain approximately 5.85 billion image-text pairs, surpassing the former by roughly \emph{three orders of magnitude}. To bridge this gap, this work proposes a novel paradigm that enables instruction-driven image editing using text-image pairs instead of image editing pairs, significantly broadening the accessibility of this task.
% This disparity in data availability is a key factor behind the robust performance of vision-language models such as CLIP\cite{CLIP}, and more recently, multimodal large language models (MLLMs) in vision-language tasks \cite{liu2023visual, liu2024improved, bai2025qwen25vltechnicalreport}. 
% \shen{So what? How does this section connect with your work?Explictly say it}
% The abundance of text-image pairs provides these models with a rich learning signal, enabling them to excel in various cross-modal understanding and generation tasks.

In this paper, we propose a novel method that leverages a pre-trained text-to-image generative model to achieve instruction-driven image editing without the need for editing-pair datasets or retraining/fine-tuning the generative model, by introducing multi-scale learnable regions. 
% Figure~\ref{fig:architecture} shows the overall pipeline of our proposed method. 
For training this model, the dataset only requires the images and the corresponding instruction prompts. Descriptions of the original and edited images are then automatically generated by MLLM and LLM, respectively, forming the dataset used in our approach and thereby expanding the accessibility of text-driven image editing.
By integrating the vision-language capabilities of CLIP, we enforce the fusion of image features and instruction features to align with the target semantic information. This fusion guides the prediction of learnable editing regions through a text-driven editing loss under CLIP supervision. The mechanism for generating these learnable regions ensures proper editing scales and accurately localized editing areas, enabling the model to handle various editing operations effectively.
Comprehensive experiments demonstrate that our method achieves state-of-the-art performance in instruction-driven image editing, delivering high realism, precise editing, various editing operations, and strong consistency in preserving unedited regions, while eliminating the dependency on instruction-editing datasets.

In summary, our contributions to the community are as follows:

\begin{enumerate}[leftmargin=10pt]

%\item We provide a systematic review of current text-driven image editing methods, categorizing them into two paradigms, and analyzing their strengths, limitations, and bottlenecks, which we hope will inspire future research in the community.
    
\item We propose a novel approach that achieves instruction-driven image editing without relying on edited-pair datasets, thus reducing data dependency and enhancing the accessibility of this task.
    
\item By introducing a learnable editing region mechanism, our method effectively handles a wide range of editing operations and accommodates different editing scales and shapes, ensuring flexible and precise control over the editing process. Moreover, the proposed method is compatible with various generative models and requires no retraining or fine-tuning.
    
\item Comprehensive experiments demonstrate that our method achieves state-of-the-art performance across various metrics and benchmarks, validating its effectiveness and generalizability.

\end{enumerate}

\section{Related Work}

\subsection{Description-driven image editing}
To achieve accurate localization of edits, description-driven approaches often require additional inputs, such as masks that specify the regions to edit, potentially limiting their practicality \cite{Goel_2024_CVPR}. Although some recent studies have addressed this limitation by introducing mask-free local editing approaches, these solutions either involve overly complex textual prompts\cite{hertz2022prompttopromptimageeditingcross, couairon2022diffeditdiffusionbasedsemanticimage, huang2024paralleledits, Cao_2023_ICCV} or roughly identify editing regions, restricting their ability to handle edits at varied scales effectively \cite{Lin_2024_CVPR}.

\subsection{Instruction-driven image editing}

% Generative models have shown impressive capabilities in generating high-quality, realistic images with strong controllability, which has also advanced text-driven image editing \cite{rombach2022high, peebles2023scalablediffusionmodelstransformers, tian2024visual}.
Methods like InstructPix2Pix~\cite{Brooks_2023_CVPR} typically combine large language models (LLMs) with data synthesis pipelines to create paired editing datasets for training~\cite{Guo_2024_CVPR, NEURIPS2023_64008fa3, NEURIPS2024_05a30a0f, hui2024hq}. % These datasets train generative models to follow textual instructions for image editing. 
However, such approaches heavily depend on large-scale, high-quality data and often result in unintended changes or artifacts in fine image details. To improve precision, recent work has introduced attention mechanisms and feature injection techniques to guide edits to specific image regions~\cite{Cao_2023_ICCV, Guo_2024_CVPR, huang2024paralleledits, Brack_2024_CVPR}. While effective for localization, these methods involve complex architectures and place constraints on data, reducing flexibility and generalization. Other strategies use segmentation to extract editing targets, enabling object-aware edits~\cite{Liu_2024_CVPR, Li_2024_CVPR, couairon2022diffeditdiffusionbasedsemanticimage, Goel_2024_CVPR, Bodur_2024_CVPR}, but are often limited to specific editing types. 
% More recent methods based on learnable regions~\cite{Lin_2024_CVPR}, which use CLIP-based loss functions, bypass the need for paired datasets and support description-driven editing. Nevertheless, they struggle with fine-grained control, varying edit scales, and complex instructions, limiting their applicability in real-world, instruction-driven scenarios. 
% \shen{I rephrased and shortened this section, read and see if it conveys your meanings}

\subsection{Multimodal large language model}
% \shen{also this section}
%After CLIP~\cite{CLIP} bridged the gap between images and text, enabling effective cross-modal understanding and retrieval, research on multimodal large language models (MLLMs) has advanced rapidly~\cite{liu2023visual, liu2024improved, bai2025qwen25vltechnicalreport}. These models extend the capabilities of traditional LLMs by incorporating visual information, allowing them to process and generate content based on both textual and visual inputs. MLLMs have demonstrated remarkable performance in a wide range of vision-language tasks, including image captioning, visual question answering, and visual reasoning \cite{liu2023visual, bai2025qwen25vltechnicalreport, fang2025got, sheynin2024emu}.

The introduction of CLIP~\cite{CLIP} marked a major breakthrough in bridging the gap between images and text, enabling effective cross-modal understanding and retrieval. Building on this foundation, research on multimodal large language models (MLLMs) has progressed rapidly~\cite{liu2023visual, liu2024improved, bai2025qwen25vltechnicalreport}. These models enhance traditional LLMs by integrating visual inputs, allowing them to jointly process and reason over both text and images. MLLMs have achieved impressive performance across a broad range of tasks, such as image captioning, visual question answering, and visual reasoning~\cite{liu2023visual, bai2025qwen25vltechnicalreport, fang2025got, sheynin2024emu}.

% While Stable diffusion successfully generate realistic and high quality image and exhibit high controllability, comes for high quality text-driven image editing. 以InstructPix2Pix\cite{Brooks_2023_CVPR}为代表的方法 通常 combines LLMs and well-designed data synthsis pipeline to generate editing pairs dataset and train a generative model using instruction as condition to achieve text-driven image editing. These method rely on large high-quality datasets and suffers from undesigned edit and high-frequency distort. To achieve more accurate and fine-grained image editing, recent works concentrate on utilize attention mechanisms and feature injection to enforce models focus on editing specific part of image, 带来了极其复杂的设计和对数据的特殊要求. Furthermore, some methods directly extract out editing object using some segmentation strategies to realize object-aware editing, which is limited to a few editing operations and flexibility. In contrast, Learnable Regions method through CLIP-based loss, achieves description-driven image editing without editing pairs datasets, however, failed in instruction fellow and various editing scale and shape. 

\section{Method}

% Use figure* for multi-column figure
\begin{figure*}[tp]
    \centering
    \includegraphics[width=1\linewidth]{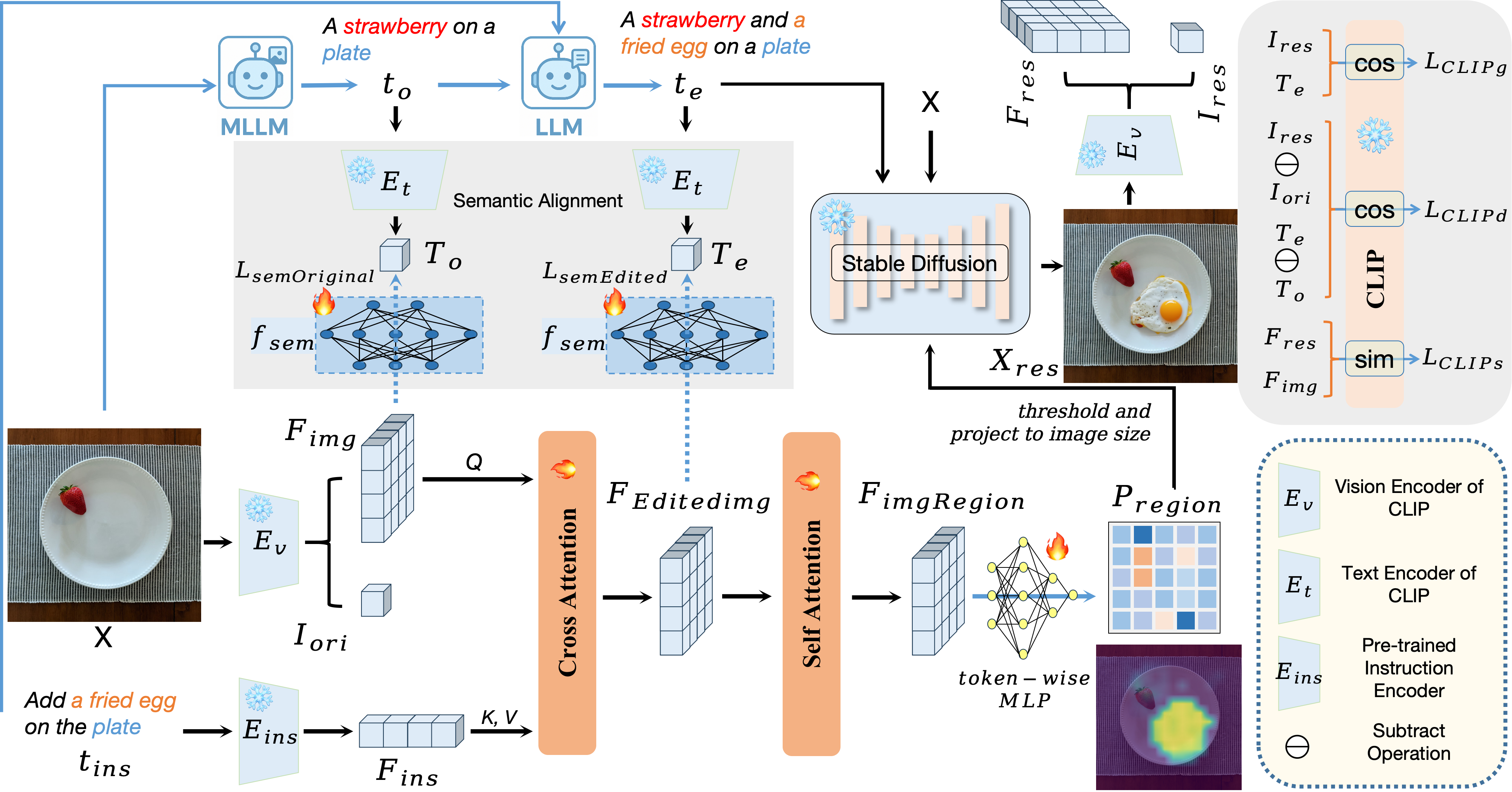}
    \caption{
    \textbf{Framework of the proposed method}.
    Including description text generation, editing feature semantic alignment, learnable edit region prediction, edited image generation and CLIP supervised loss calculation.
    \includegraphics[height=0.8em]{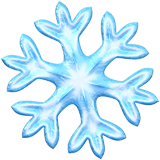} means the parameters of the component remain fixed, and
    \includegraphics[height=0.8em]{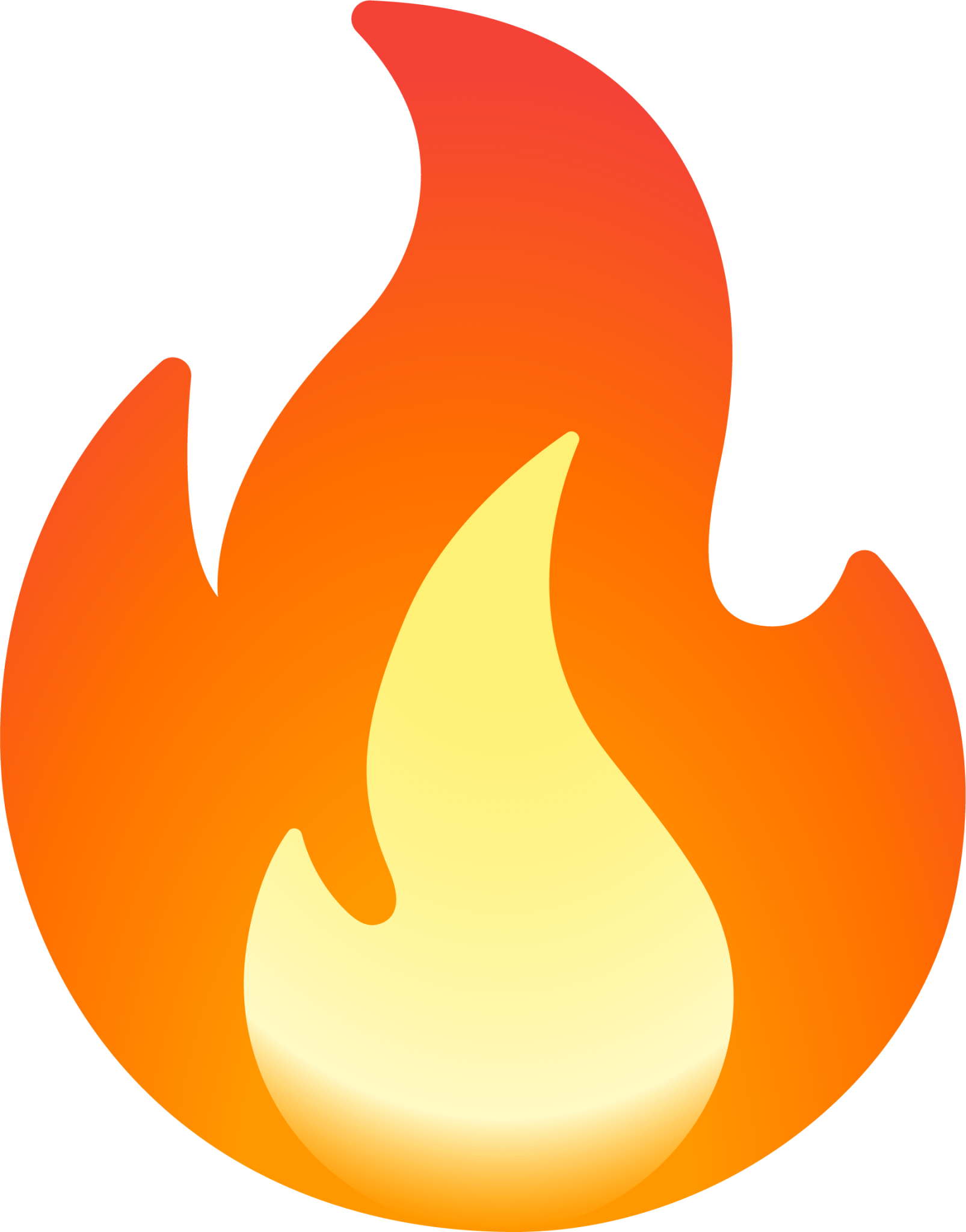} means the parameters of the component are activated for training.
    During the inference stage, the components in the gray area will be removed.
    Stable Diffusion \cite{rombach2022high} serves as the generative model here; however, various text-to-image generators can be chosen, refer to \hyperref[ImplementDetails]{Implementation Details} for more information.
    }
    \vspace{-1em}
\label{fig:architecture}
\end{figure*}

Our proposed method enables instruction-driven image editing via flexible learnable regions, while leveraging pre-trained models, without requiring paired editing data or model fine-tuning. 
The core idea is to automatically generate textual descriptions, utilize the joint embedding space of CLIP to guide feature fusion and region identification, and employ a pre-trained text-to-image generative model conditioned on the learnable region for image synthesis. 
See our framework in Figure~\ref{fig:architecture}.
% The overall process involves descriptive text generation, feature extraction, instruction-guided feature fusion, learnable edit region prediction, guided image generation, and a composite loss function for optimization.

\subsection{Descriptive Text Generation and Feature Preparation}

Given an input image $X$ and a textual editing instruction $t_{\text{ins}}$ (e.g., ``change the cat into a dog''), we first automate the generation of relevant textual descriptions. To this end, we employ a Multimodal Large Language Model (MLLM) to generate a description of the original image, denoted as $t_o$ (e.g., ``a cat''): $t_o = \text{MLLM}(X)$. Subsequently, we utilize a Large Language Model (LLM), providing it with both the original description $t_o$ and the editing instruction $t_{\text{ins}}$, to synthesize a description of the target edited image, denoted as $t_e$ (e.g., ``a dog''): $t_e = \text{LLM}(t_o, t_{ins})$. 
% This automated process eliminates the need for manually curated datasets. 
It simply involves image captioning and text generation or conversion, which current models can perform very well. This approach completely avoids the complex, time-consuming, and low-quality process of creating editing pairs.

We then extract essential feature representations using the text encoder $E_t$ of the CLIP model to obtain global text embeddings (CLS tokens) for the original and target descriptions:
$T_o = E_t(t_o)$, $T_e = E_t(t_e)$.
% \begin{equation}
%     T_o = E_t(t_o)
%     \label{eq:def_To}
% \end{equation}
% \begin{equation}
%     T_e = E_t(t_e)
%     \label{eq:def_Te}
% \end{equation}
These embeddings, $T_o, T_e \in \mathbb{R}^d$, capture the semantic essence of the descriptions in the CLIP space aligned with visual concepts, where $d$ is the CLIP embedding dimension.

Similarly, we employ the visual encoder $E_v$ of CLIP to extract features from the source image $X$. We obtain the global image embedding (CLS token) $I_{\text{ori}} \in \mathbb{R}^d$, which aligned with textual concept~\cite{CLIP}: $I_{\text{ori}} = E_v(X)^{\text{[CLS]}}$.
% \begin{equation}
%     I_{\text{ori}} = E_v(X)_{\text{[CLS]}}
%     \label{eq:def_Iori}
% \end{equation}
Furthermore, we extract the patch-level features that contain spatial information from the visual encoder, denoted as $F_{\text{img}} \in \mathbb{R}^{N \times d}$, where $N$ is the number of visual tokens (patches) and $d$ is the feature dimension per token: $F_{\text{img}} = E_v(X)^{\text{[patches]}}$.
% \begin{equation}
%     F_{\text{img}} = E_v(X)_{\text{[patches]}}
%     \label{eq:def_Fimg}
% \end{equation}
Finally, the editing instruction $t_{\text{ins}}$ is encoded using a pre-trained text encoder, denoted as $E_{\text{ins}}$, to obtain instruction features $F_{\text{ins}} \in \mathbb{R}^{M \times d}$, where $M$ is the sequence length of the instruction tokens and $d$ is the corresponding feature dimension, which is specially designed to equal to the feature dimension of CLIP encoders for simplicity and effective fusion: $F_{\text{ins}} = E_{\text{ins}}(t_{\text{ins}})$
% \begin{equation}
%     F_{\text{ins}} = E_{\text{ins}}(t_{\text{ins}})
%     \label{eq:def_Fins}
% \end{equation}

\subsection{Learnable Edit Region Prediction and Instruction-followed Image Editing}

To effectively leverage both the visual content and the editing intent, we fuse information from the source image $X$ with guidance from the editing instruction $t_{ins}$, directing the editing process toward the most relevant regions. This is achieved through the use of attention mechanisms.

First, we perform cross-attention between the image features $F_{\text{img}}$ and the instruction features $F_{\text{ins}}$. Treating $F_{\text{img}}$ as the query ($Q$) sequence and $F_{\text{ins}}$ as the key ($K$) and value ($V$) sequences, we compute the instruction-fused image features $F_{\text{Editedimg}} \in \mathbb{R}^{N \times d}$:
% \begin{equation}
%     F_{\text{Editedimg}} = \text{CrossAttn}(Q=F_{\text{img}}, K=F_{\text{ins}}, V=F_{\text{ins}})
%     \label{eq:calc_Feditedimg}
% \end{equation}
\begin{equation}
\begin{aligned}
Q^{C} = F_{\text{img}} W_Q^{C}, \quad
&K^{C} = F_{\text{ins}} W_K^{C}, \quad
V^{C} = F_{\text{ins}} W_V^{C}, \quad\quad\quad  W_Q^{C}, W_K^{C}, W_V^{C} \in \mathbb{R}^{d \times d}
\\
&F_{\text{Editedimg}} = \text{Softmax}\left(\frac{Q^{C} {K^{C}}^\top}{\sqrt{d}}\right)V^{C} \in \mathbb{R}^{N \times d}
\end{aligned}
\label{eq:calc_Feditedimg}
\end{equation}
This operation allows each image patch feature to attend to the instruction features, incorporating editing guidance into the visual representation. The resulting $F_{\text{Editedimg}}$ ideally encodes spatial information modulated by the edit instruction.

To further refine the spatial localization of the edit and capture contextual relationships within the potentially edited regions, we apply a self-attention mechanism to the fused features:
% \begin{equation}
%     F_{\text{imgRegion}} = \text{SelfAttn}(F_{\text{Editedimg}})
%     \label{eq:calc_FimgRegionPrime}
% \end{equation}
\begin{equation}
\begin{aligned}
Q^{S} = F_{\text{Editedimg}} W_Q^{S}, \ \
&K^{S} = F_{\text{Editedimg}} W_K^{S}, \ \
V^{S} = F_{\text{Editedimg}} W_V^{S}, \quad \ W_Q^{S}, W_K^{S}, W_V^{S} \in \mathbb{R}^{d \times d}
\\
&F_{\text{imgRegion}} = \text{Softmax}\left(\frac{Q^{S} {K^{S}}^\top}{\sqrt{d}}\right)V^{S} \in \mathbb{R}^{N \times d}
\end{aligned}
\label{eq:calc_FimgRegionPrime}
\end{equation}
% where $F_{\text{imgRegion}} \in \mathbb{R}^{N \times d}$.
Here $F_{\text{imgRegion}}$ integrates both visual and instruction-driven semantics in a context-aware manner.
Based on these refined features $F_{\text{imgRegion}}$, we predict a spatial mask indicating the regions designated for editing. We employ a simple Multi-Layer Perceptron (MLP) applied token-wise (i.e., per patch feature) to $F_{\text{imgRegion}}$ and followed by Sigmoid operation to generate a probability map $P_{\text{region}} \in \mathbb{R}^{N}$:
\begin{equation}
    P_{\text{region}} = \text{Sigmoid}(\text{MLP}(F_{\text{imgRegion}}))
    \label{eq:calc_Pregion}
\end{equation}
Each element in $P_{\text{region}}$ corresponds to a patch in the original image feature map $F_{\text{img}}$, representing the likelihood that this patch belongs to the edit region. This probability map is then thresholded and reshaped to a 2D grid corresponding to the image patch layout to form the final editing region mask, denoted as $M_{\text{region}}$. This learnable mask $M_{\text{region}}$ guides the subsequent image generation process, ensuring that edits are localized and appropriately scaled.

We utilize a pre-trained text-to-image generative model for generating the final edited image $X_{\text{res}}$. Note that our method is compatible with various generative models as detailed in \hyperref[ImplementDetails]{Implementation Details} and \hyperref[Appendix_A]{Appendix A}. Crucially, we do not fine-tune the generative model itself. Instead, we condition its generation process using the original image $X$, the target text description $t_e$, and our predicted editing region mask $M_{\text{region}}$, 
% The specific mechanism involves incorporating $M_{\text{region}}$ into the diffusion process (e.g., by using it to blend latent representations or guide the denoising steps), 
ensuring that the generative process primarily modifies the areas identified by $M_{\text{region}}$ while preserving the content in unmasked regions, guided by the target description $t_e$. The resulting edited image is denoted as $X_{\text{res}}$.

\subsection{Optimization Objective}

To train the components responsible for feature fusion and learnable edit region prediction, specifically, the parameters of the cross-attention~\eqref{eq:calc_Feditedimg}, self-attention~\eqref{eq:calc_FimgRegionPrime} modules, the MLP~\eqref{eq:calc_Pregion} for region prediction, and the semantic alignment network $f_{\text{sem}}$ introduced below, we define a composite loss function consisting of two main parts: a semantic alignment loss $L_{\text{semAlign}}$ and a CLIP supervision loss $L_{\text{CLIP}}$.

\textbf{Semantic Alignment Loss ($L_{\text{semAlign}}$):} This loss ensures that the fused feature representation $F_{\text{Editedimg}}$ effectively captures the semantics of the target description $T_e$. We introduce a learnable network $f_{\text{sem}}$ that projects the spatial features $F_{\text{Editedimg}}$ to the CLIP embedding space. We then compute the cosine similarity between the predicted embedding and the target text embedding $T_e$. To prevent the network $f_{\text{sem}}$ from collapsing to trivial solutions (e.g., overfitting to $F_{\text{Editedimg}}$), we also enforce semantic alignment for the original image features $F_{\text{img}}$ with the original description embedding $T_o$ using the same network $f_{\text{sem}}$. The losses are defined using cosine distance ($1 - \cos(\cdot, \cdot)$):
\begin{equation}
    L_{\text{semEdited}} = 1 - \cos(f_{\text{sem}}(F_{\text{Editedimg}}), T_e)
    \label{eq:loss_semEdited}
\end{equation}
\begin{equation}
    L_{\text{semOriginal}} = 1 - \cos(f_{\text{sem}}(F_{\text{img}}), T_o)
    \label{eq:loss_semOriginal}
\end{equation}
The total semantic alignment loss is the sum of \eqref{eq:loss_semEdited},\eqref{eq:loss_semOriginal}:
\begin{equation}
    L_{\text{semAlign}} = L_{\text{semEdited}} + L_{\text{semOriginal}}
    \label{eq:loss_semAlign}
\end{equation}
This encourages spatial feature $F_{\text{Editedimg}}$ to be semantically aligned with the desired edit outcome.

\textbf{CLIP Supervision Loss ($L_{\text{CLIP}}$):} This loss leverages CLIP to guide the generation of the final output image $X_{\text{res}}$. In parallel with the spatial-level supervision provided by $L_{\text{semAlign}}$~\eqref{eq:loss_semAlign}, it offers additional global guidance toward the target semantics\cite{Liu_2024_CVPR, patashnik2021styleclip, crowson2022vqgan}. It consists of three components:

\begin{enumerate}[leftmargin=10pt]
    \item \textbf{CLIP Guidance Loss ($L_{\text{CLIPg}}$):} This loss encourages the generated image $X_{\text{res}}$ to be semantically aligned with the target description $t_e$. We extract the global image embedding as $I_{\text{res}} = E_v(X_{\text{res}})^{\text{[CLS]}}$, and compute the cosine distance between $I_{\text{res}}$ and the target text embedding $T_e$:

    % {\color{blue}what is the [CLS], do you really need it? if removing it does not lead to any ambiguity, I suggest you remove it}

    \begin{equation}
        L_{\text{CLIPg}} = 1 - \cos(I_{\text{res}}, T_e)
        \label{eq:loss_CLIPg}
    \end{equation}

    \item \textbf{Directional CLIP Loss ($L_{\text{CLIPd}}$):} This loss guides the editing process in the CLIP embedding space~\cite{Liu_2024_CVPR} by aligning the direction of change in the image domain with that in the text domain. Specifically, it encourages the vector difference between the original image embedding $I_{\text{ori}}$ and the edited image embedding $I_{\text{res}}$ to align with the difference between the original text embedding $T_o$ and the target text embedding $T_e$:

    \begin{equation}
        L_{\text{CLIPd}} = 1 - \cos(I_{\text{res}} - I_{\text{ori}}, T_e - T_o)
        \label{eq:loss_CLIPd}
    \end{equation}

    \item \textbf{Structural Similarity Loss ($L_{\text{CLIPs}}$):} This loss is designed to preserve the spatial layout and structure of the original image~\cite{patashnik2021styleclip}. We extract thce patch-level features $F_{\text{res}} = E_v(X_{\text{res}})^{\text{[patches]}}$ for the generated image, then compute the similarity matrix (i.e., pairwise cosine similarities between patch features), for both the original features $F_{\text{img}}$ and the generated features $F_{\text{res}}$ respectively. The loss is the L2 distance between these similarity matrices:

    \begin{equation}
        L_{\text{CLIPs}} = || \text{Sim}(F_{\text{img}}) - \text{Sim}(F_{\text{res}}) ||_2^2, \text{  where }  \text{Sim}(F) = \left[ \frac{F_i \cdot F_j}{||F_i|| \, ||F_j||} \right]_{i,j}
        \label{eq:loss_CLIPs}
    \end{equation}
    where $F_i$ and $F_j$ are the feature vectors for the i-th and j-th image patches, respectively, $F_i \cdot F_j$ is their dot product, and $||F_i||$, $||F_j||$ are their respective L2 norms.
    % This encourages the preservation of relative spatial relationships between elements in the image.
\end{enumerate}

The overall CLIP supervision loss is a weighted sum of these components \eqref{eq:loss_CLIPg},\eqref{eq:loss_CLIPd},\eqref{eq:loss_CLIPs}:
\begin{equation}
    L_{\text{CLIP}} = \lambda_{g} L_{\text{CLIPg}} + \lambda_{d} L_{\text{CLIPd}} + \lambda_{s} L_{\text{CLIPs}}
    \label{eq:loss_CLIP}
\end{equation}
% \shen{remove one of the $\lambda$}
where $\lambda_{g}, \lambda_{d}, \lambda_{s}$ are hyperparameters balancing the contributions.

\textbf{Overall Loss:} The final objective function optimized during training is a weighted combination of the semantic alignment~\eqref{eq:loss_semAlign} and CLIP supervision~\eqref{eq:loss_CLIP} losses:
\begin{equation}
    L_{\text{total}} = \alpha L_{\text{semAlign}} + \beta L_{\text{CLIP}}
    \label{eq:loss_Total}
\end{equation}
% \shen{remove $\alpha$ or $\beta$}
where $\alpha$ and $\beta$ are hyperparameters controlling the influence of each major loss component. By minimizing $L_{\text{total}}$, we train the fusion and learnable region modules to effectively guide the pre-trained text-to-image generative model for precise and instruction-aligned image editing. Notably, both the text-to-image generative model and the CLIP encoder remain fixed during training.

\section{Implementation Details} \label{ImplementDetails}

Our method is compatible with various text-to-image generative models, enabling instruction-driven image editing without the need for editing pair datasets or fine-tuning the generative models, as these generative models are able to perform generation process conditioned on text and generating region. We consider three distinct categories of generative models that represent the major paradigms in current research: diffusion models~\cite{rombach2022high, flux2024}, autoregressive generative models~\cite{tian2024visual, peebles2023scalablediffusionmodelstransformers}, and non-autoregressive generative models~\cite{liu2021pd}. For fair comparison and experimental consistency, we adopt the diffusion model as our backbone~\cite{podell2023sdxlimprovinglatentdiffusion}, as shown in \hyperref[Appendix_A]{Appendix A}, since it currently represents the state-of-the-art and is widely used in text-to-image generation. Nevertheless, we also demonstrate that our method is adaptable to other generative model architectures, highlighting its broad applicability, as shown in \hyperref[compatibility]{Experiment}.

In our experiment setup, we choose the ViT-L/14 version of CLIP \cite{CLIP}, whose token dimension is 768, for our framework. To encode the editing instruction, we utilize FLAN-T5-Base \cite{chung2022scaling} as our pre-trained text encoder $E_{\text{ins}}$, ensuring its token dimension also equals 768, consistent with CLIP's output dimension. Regarding the choice of the LLM and MLLM, Many leading models were qualified based on criteria including benchmark performance, architectural suitability for our task~\cite{openai2024gpt4o, anthropic2024claude3.5, google2025gemini25pro, meta2024llama3.1, liu2024deepseek}, we selected Qwen2.5 \cite{qwen_team2024qwen2.5} and Qwen2.5-VL \cite{qwen_team2025qwen2.5vl}, respectively. Both models represent state-of-the-art performance within the open-source domain at the time of selection and align well with our framework's requirements.

We train our model~\eqref{eq:calc_Feditedimg}\eqref{eq:calc_FimgRegionPrime}\eqref{eq:calc_Pregion} using text-image pair datasets, which are significantly larger than typical editing pair datasets, as discussed in the \hyperref[intro]{Introduction}. Given the large scale of these datasets (approximately 400M pairs~\cite{schuhmann2021laion} or even 5B pairs~\cite{schuhmann2022laion}), we randomly sample a subset of 5 million pairs for training. 
Each data sample in the dataset contains an original image $X$ and its associated description $t_o$. To construct the required training data, we employ the LLM~\cite{qwen_team2024qwen2.5} to automatically generate a suitable editing instruction $t_{\text{ins}}$ and a corresponding target description $t_e$ based on $t_o$ and $t_{\text{ins}}$.

As illustrated in Figure~\ref{fig:architecture}, during inference, the user provides an input image $X$ and an editing instruction $t_{\text{ins}}$. The MLLM~\cite{qwen_team2025qwen2.5vl} generates the original description $X_o$, while the LLM~\cite{qwen_team2024qwen2.5} produces the target description $t_e$ to guide the editing process.

\begin{figure}[htbp] % Begin figure environment
    \centering % Center the figure content horizontally

    % --- Row of 7 Images ---
    % Use [t] alignment for all subfigures to ensure images align at the top
    % Calculate width: 1/7 is approx 0.14. Use slightly less for spacing.
    % Let's try 0.135 * 7 = 0.945, leaving space for \hfill gaps. Adjust if needed.
    
    \begin{subfigure}[t]{0.135\textwidth} % Set width for first image
        \includegraphics[width=\linewidth]{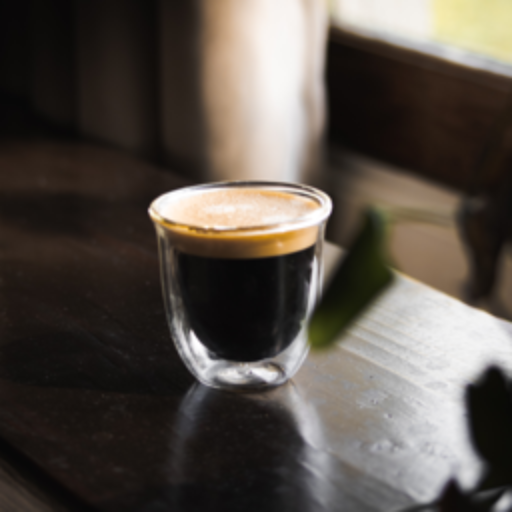} % Image file. \linewidth refers to subfigure width
        \caption*{\scriptsize A cup of coffee} % Caption for image 1
    \end{subfigure}% <-- The '%' is crucial to prevent unwanted space after the subfigure
    \hfill % Add flexible horizontal space between images
    \begin{subfigure}[t]{0.135\textwidth} % Set width for second image
        \includegraphics[width=\linewidth]{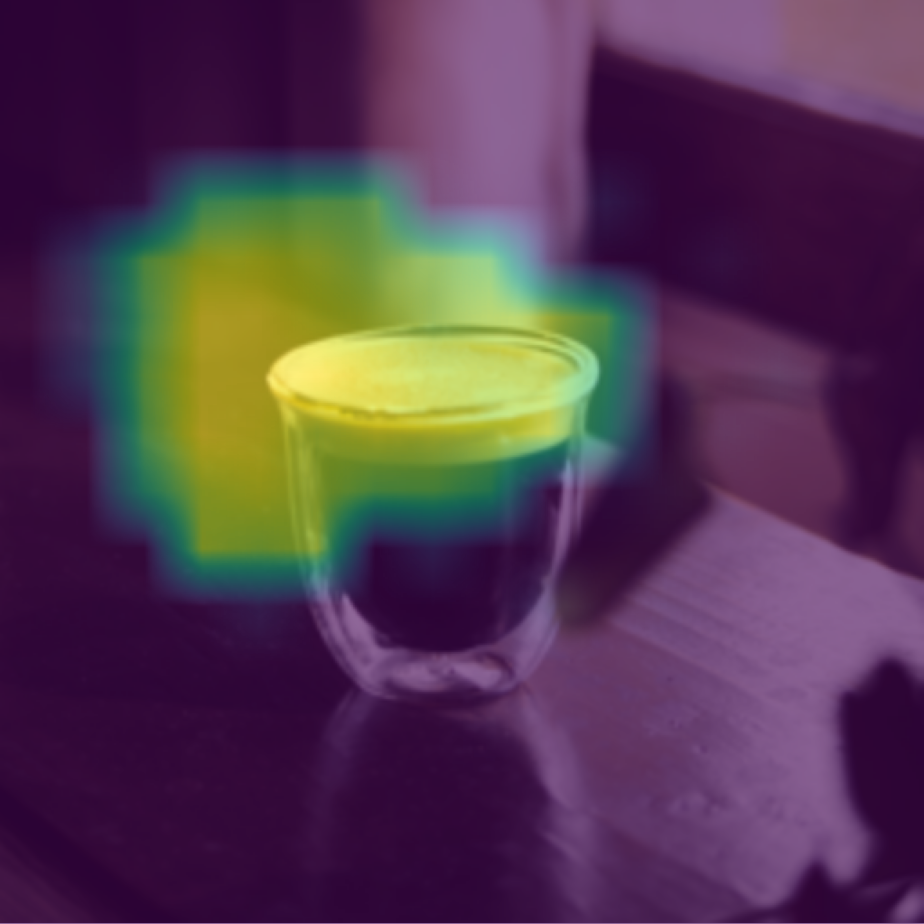} % Image file 2
        \caption*{\scriptsize Add a piece of lemon to the cup of coffee} % Caption for image 2
    \end{subfigure}%
    \hfill
    \begin{subfigure}[t]{0.135\textwidth} % Set width for third image
        \includegraphics[width=\linewidth]{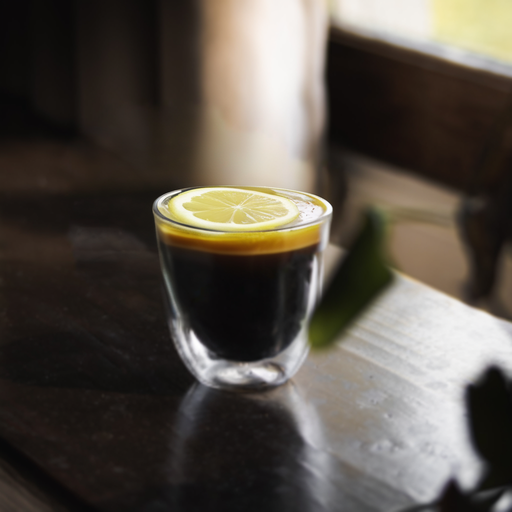} % Image file 3
        \caption*{\scriptsize A cup of coffee with a piece of lemon} % Caption for image 3
    \end{subfigure}%
    \hfill
    \begin{subfigure}[t]{0.135\textwidth} % Set width for fourth image
        \includegraphics[width=\linewidth]{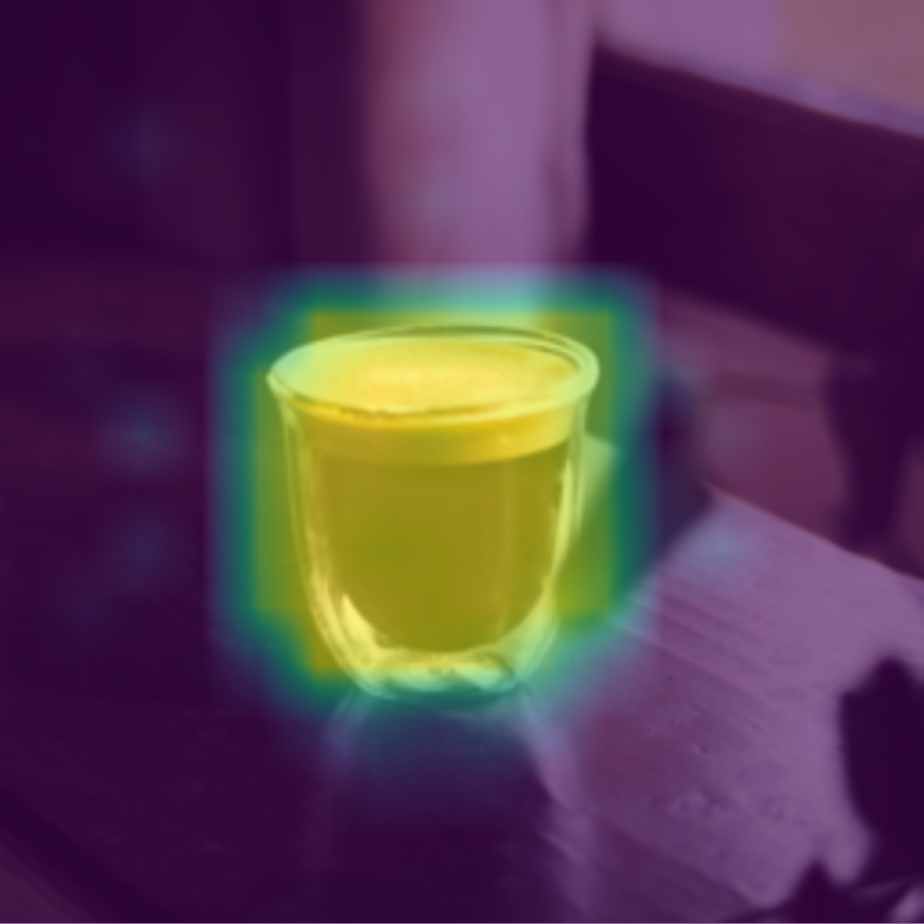} % Image file 4
        \caption*{\scriptsize Change the coffee in the cup to soft drink} % Caption for image 4
    \end{subfigure}%
    \hfill
    \begin{subfigure}[t]{0.135\textwidth} % Set width for fifth image
        \includegraphics[width=\linewidth]{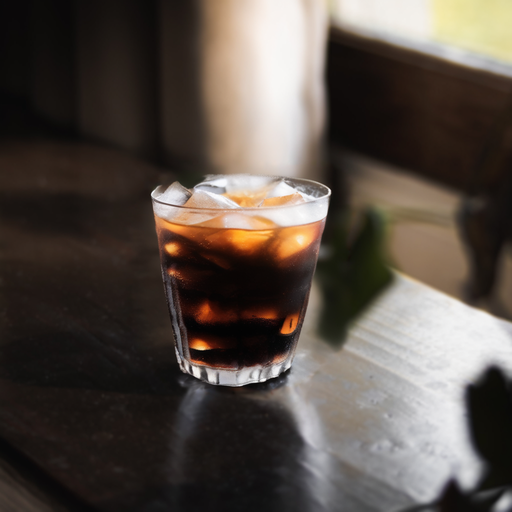} % Image file 5
        \caption*{\scriptsize A cup of soft drink} % Caption for image 5
    \end{subfigure}%
    \hfill
    \begin{subfigure}[t]{0.135\textwidth} % Set width for sixth image
        \includegraphics[width=\linewidth]{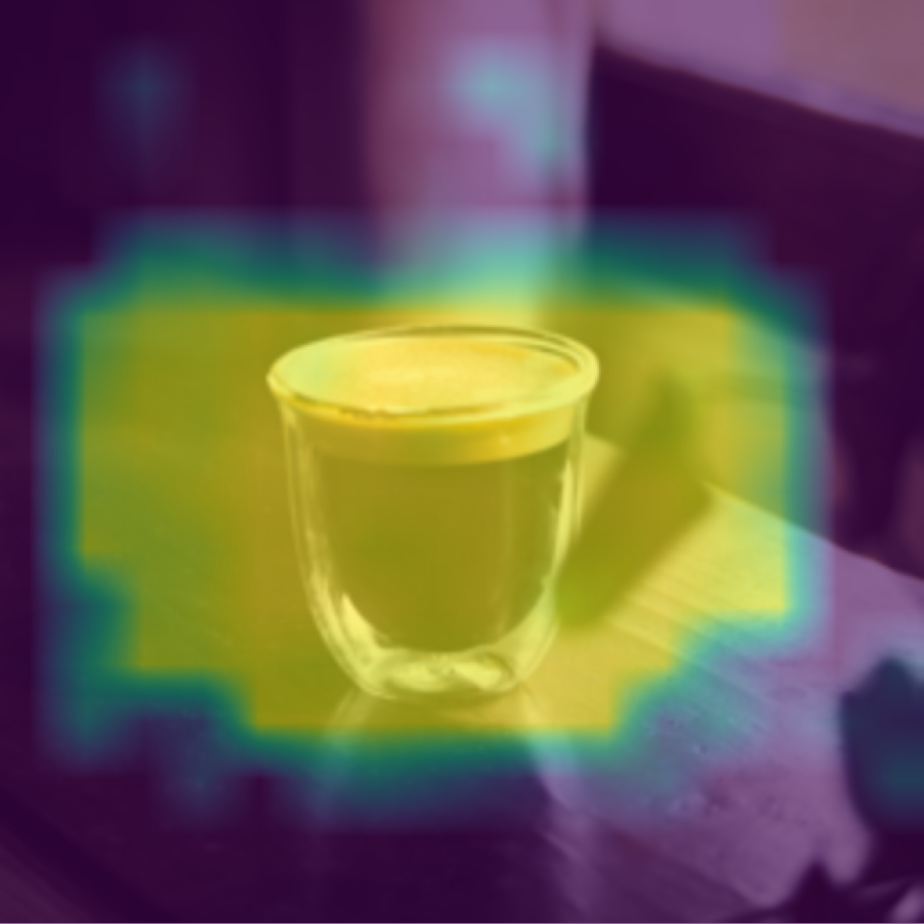} % Image file 6
        \caption*{\scriptsize Change the cup of coffee to a bowl of noodles} % Caption for image 6
    \end{subfigure}%
    \hfill
    \begin{subfigure}[t]{0.135\textwidth} % Set width for seventh image
        \includegraphics[width=\linewidth]{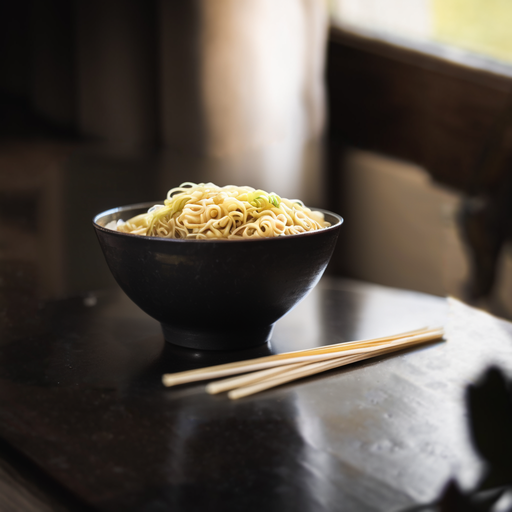} % Image file 7
        \caption*{\scriptsize A bowl of noodles} % Caption for image 7
    \end{subfigure} % No '%' needed after the last subfigure on the line

    \begin{subfigure}[t]{0.135\textwidth} % Set width for first image
        \includegraphics[width=\linewidth]{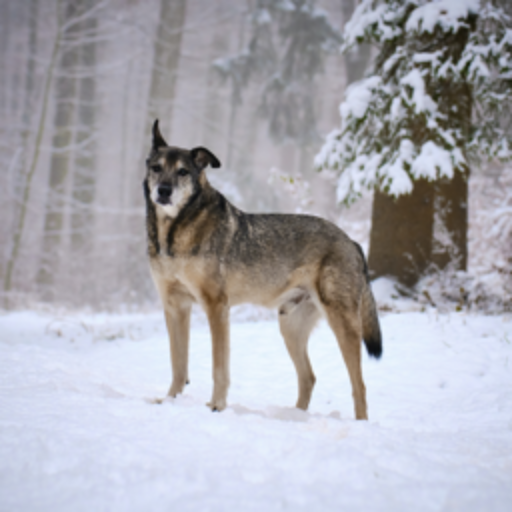} % Image file. \linewidth refers to subfigure width
        \caption*{\scriptsize A wolf} % Caption for image 1
    \end{subfigure}% <-- The '%' is crucial to prevent unwanted space after the subfigure
    \hfill % Add flexible horizontal space between images
    \begin{subfigure}[t]{0.135\textwidth} % Set width for second image
        \includegraphics[width=\linewidth]{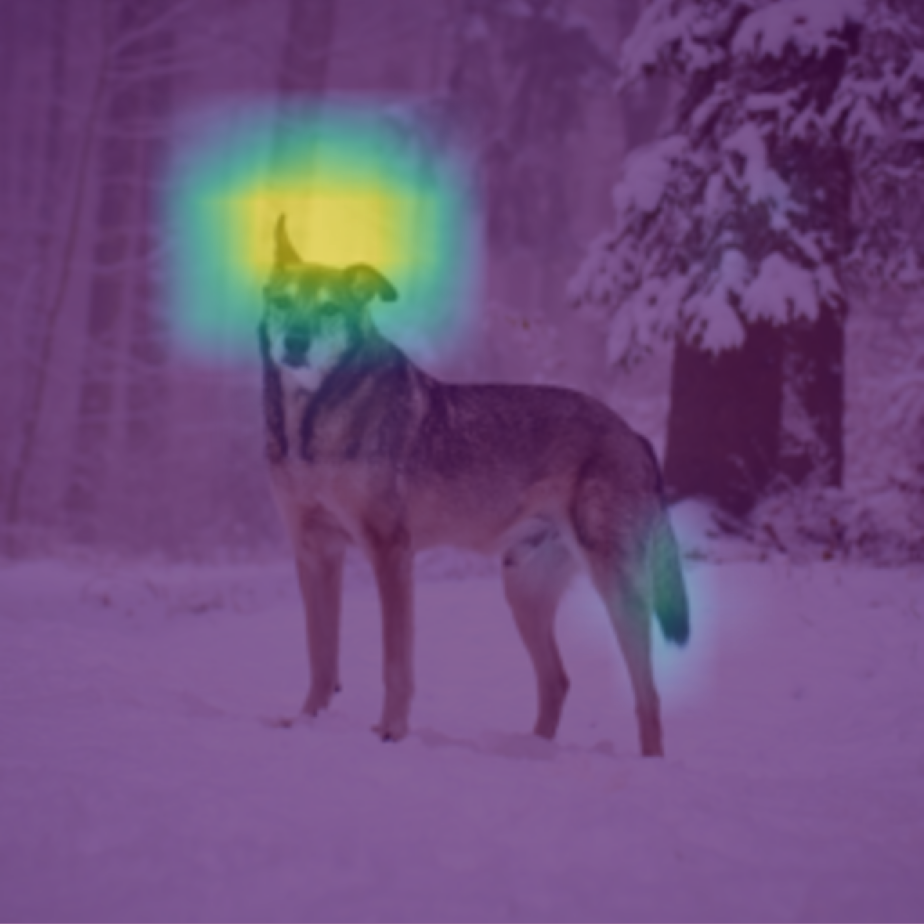} % Image file 2
        \caption*{\scriptsize Add the wolf with a golden crown} % Caption for image 2
    \end{subfigure}%
    \hfill
    \begin{subfigure}[t]{0.135\textwidth} % Set width for third image
        \includegraphics[width=\linewidth]{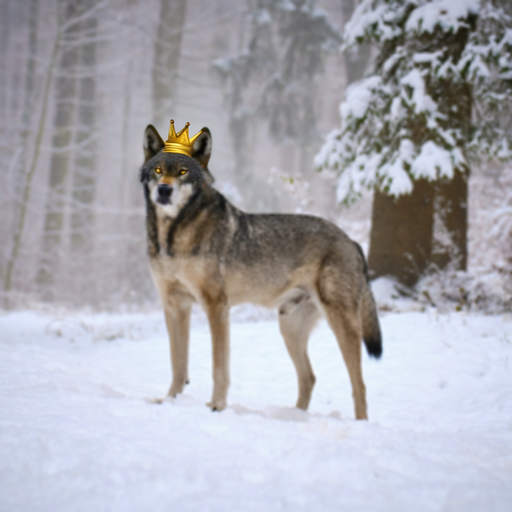} % Image file 3
        \caption*{\scriptsize A wolf with a golden crown} % Caption for image 3
    \end{subfigure}%
    \hfill
    \begin{subfigure}[t]{0.135\textwidth} % Set width for fourth image
        \includegraphics[width=\linewidth]{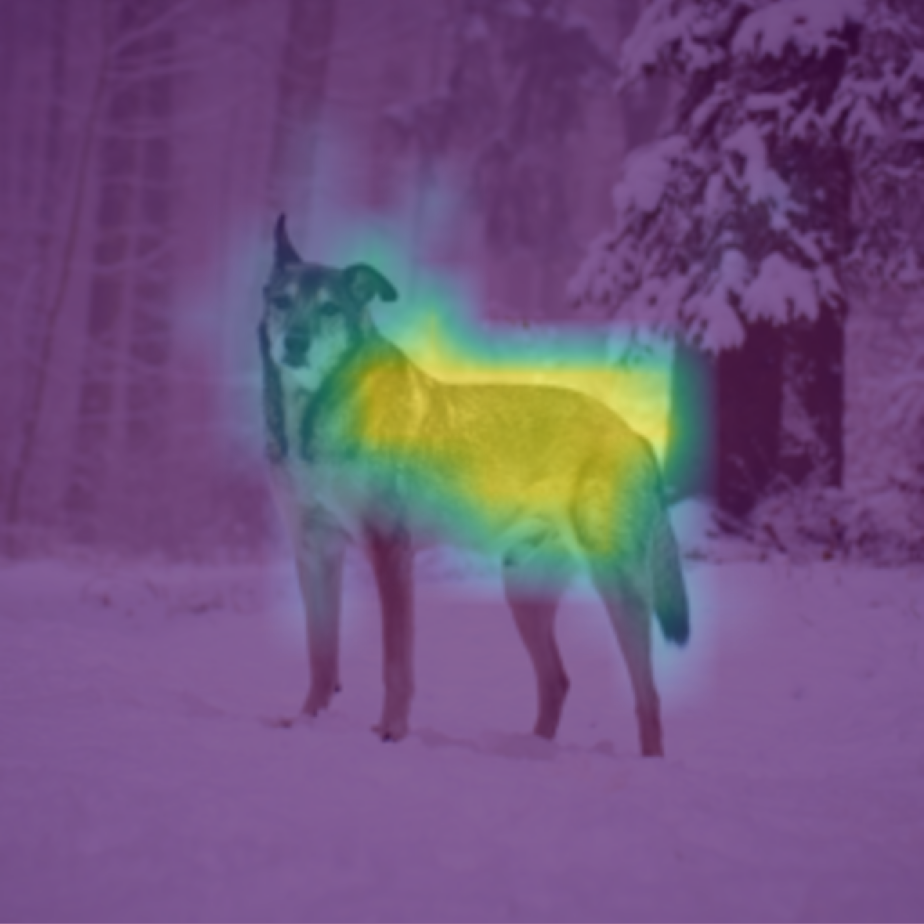} % Image file 4
        \caption*{\scriptsize Wear the wolf in a red cape} % Caption for image 4
    \end{subfigure}%
    \hfill
    \begin{subfigure}[t]{0.135\textwidth} % Set width for fifth image
        \includegraphics[width=\linewidth]{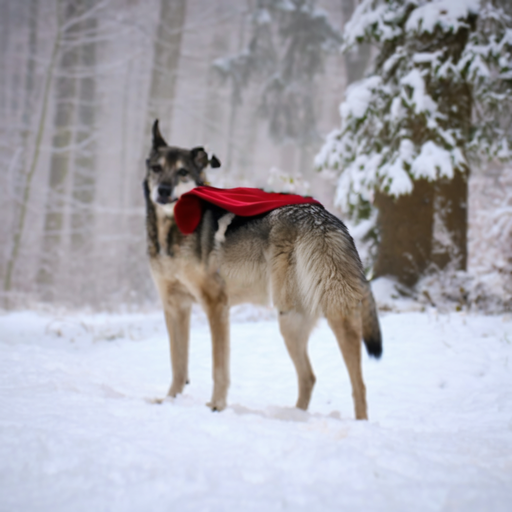} % Image file 5
        \caption*{\scriptsize A wolf wearing a red cape} % Caption for image 5
    \end{subfigure}%
    \hfill
    \begin{subfigure}[t]{0.135\textwidth} % Set width for sixth image
        \includegraphics[width=\linewidth]{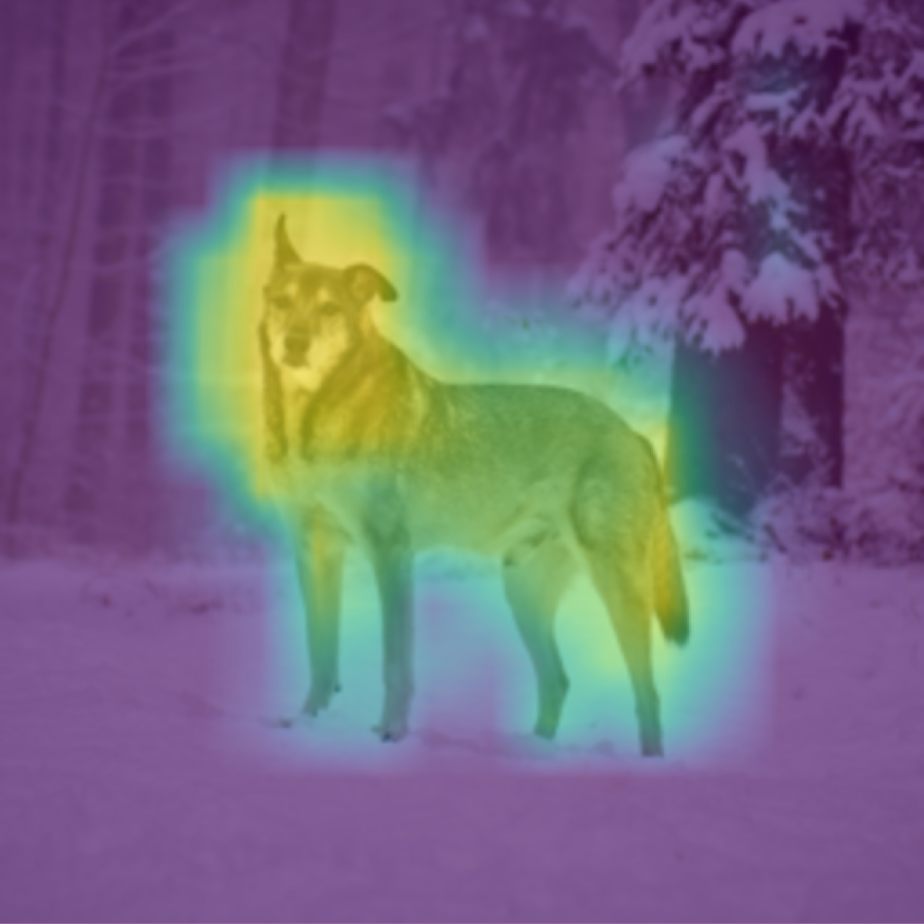} % Image file 6
        \caption*{\scriptsize Replace the wolf with a cow} % Caption for image 6
    \end{subfigure}%
    \hfill
    \begin{subfigure}[t]{0.135\textwidth} % Set width for seventh image
        \includegraphics[width=\linewidth]{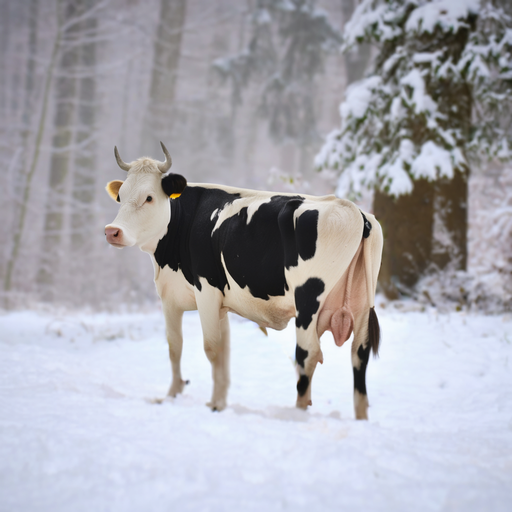} % Image file 7
        \caption*{\scriptsize a cow} % Caption for image 7
    \end{subfigure} % No '%' needed after the last subfigure on the line

    % --- Overall Caption for the Entire Figure ---
    \caption{\textbf{Multi-Scale Learnable Region.} The learnable region adapts to multi-scale editing requirements from different types of editing operations and varying sizes of target objects.}

    \label{fig:multiScale} % Label for the entire figure
    \vspace{-1em}
\end{figure} % End figure environment

\section{Experiment}
To evaluate the performance and effectiveness of our method, we conduct comprehensive experiments, including both qualitative and quantitative evaluations, as well as observations derived from experimental results and user feedback(shown in Figure~\ref{fig:preference}).

\textbf{Compared Methods:}

% 定义符号
\newcommand{\cmark}{\textcolor{green!60!black}{\checkmark}}  
\newcommand{\xmark}{\textcolor{red!80!black}{\ding{55}}}  

\begin{table}[t]
\centering
\begin{minipage}{0.7\textwidth}
\raggedright
\resizebox{\linewidth}{!}{%
\begin{tabular}{lcccc}
\toprule
\textbf{Methods} & 
\shortstack{\textbf{Instruction}\\\textbf{Followed}} & 
\shortstack{\textbf{Various}\\\textbf{Editing Ops}} & 
\shortstack{\textbf{Localized}\\\textbf{Editing}} & 
\shortstack{\textbf{Editing Pairs}\\\textbf{Data Free}} \\
\midrule
InstructPix2Pix~\cite{Brooks_2023_CVPR} & \cmark & \cmark & \xmark & \xmark \\
MagicBrush~\cite{NEURIPS2023_64008fa3} & \cmark & \cmark & \xmark & \xmark \\
UltraEdit~\cite{NEURIPS2024_05a30a0f} & \cmark & \cmark & \xmark & \xmark \\
LearnRegion~\cite{Lin_2024_CVPR} & \xmark & \xmark & \cmark & \cmark \\
RF-Solver~\cite{wang2024taming} & \xmark & \cmark & \xmark & \cmark \\
Plug-and-Play~\cite{tumanyan2023plug}  & \xmark & \xmark & \xmark & \cmark \\
\midrule
\textbf{Ours} & \cmark & \cmark & \cmark & \cmark \\
\bottomrule
\end{tabular}
}
\captionof{table}{\textbf{Comparison of Editing Capabilities.} \cmark\ indicates support for a feature; \xmark\ indicates lack of support. Our method offers high flexibility and accessibility in editing.}
\label{tab:editingCapability}

\end{minipage}%
\hfill
\begin{minipage}{0.3\textwidth}
\centering
\includegraphics[width=\linewidth]{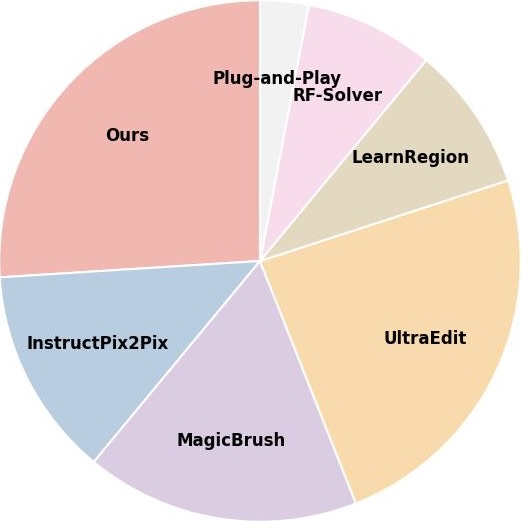}
\captionof{figure}{Illustration of user preferences for edited results.}
\label{fig:preference}
\end{minipage}
\vspace{-2em}

\end{table}

We select several state-of-the-art methods for instruction-driven image editing as our comparison baselines, including InstructPix2Pix~\cite{Brooks_2023_CVPR}, MagicBrush~\cite{NEURIPS2023_64008fa3}, and UltraEdit~\cite{NEURIPS2024_05a30a0f}. To broaden the scope of comparison, we also include the description-driven method LearnRegion~\cite{Lin_2024_CVPR}, the noise inversion-based approach RF-Solver~\cite{wang2024taming}, and the text-driven image translation method Plug-and-Play~\cite{tumanyan2023plug}. We summarize the editing capabilities of each method in Table~\ref{tab:editingCapability}, which demonstrates that our method excels in both accessibility and editing flexibility.

\subsection{Qualitative Evaluation}

\begin{figure}[htbp]
    \centering

    % ==== Column headers ====
    \begin{minipage}[t]{0.119\textwidth}
        \centering
        \scriptsize Original Image
    \end{minipage}
    \hfill
    \begin{minipage}[t]{0.119\textwidth}
        \centering
        \scriptsize InstructPix2Pix
    \end{minipage}
    \hfill
    \begin{minipage}[t]{0.119\textwidth}
        \centering
        \scriptsize MagicBrush
    \end{minipage}
    \hfill
    \begin{minipage}[t]{0.119\textwidth}
        \centering
        \scriptsize UltraEdit
    \end{minipage}
    \hfill
    \begin{minipage}[t]{0.119\textwidth}
        \centering
        \scriptsize LearnRegion
    \end{minipage}
    \hfill
    \begin{minipage}[t]{0.119\textwidth}
        \centering
        \scriptsize RF-Solver
    \end{minipage}
    \hfill
    \begin{minipage}[t]{0.119\textwidth}
        \centering
        \scriptsize Plug-and-Play
    \end{minipage}
    \hfill
    \begin{minipage}[t]{0.119\textwidth}
        \centering
        \scriptsize Ours
    \end{minipage}

    % \vspace{0.5em}

    % ==== Row 1: 8 images ====
    \begin{minipage}[t]{0.119\textwidth}
        \includegraphics[width=\linewidth]{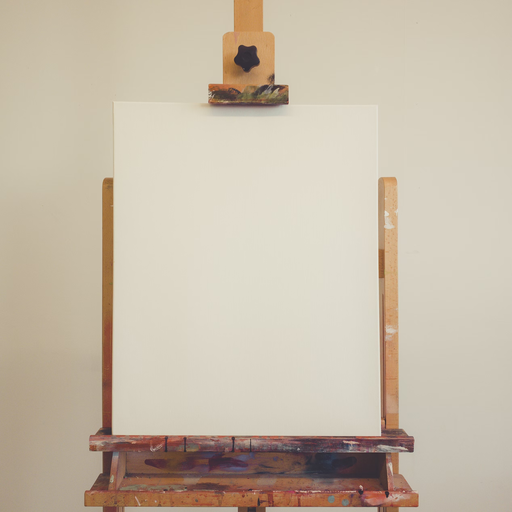}
    \end{minipage}
    \hfill
    \begin{minipage}[t]{0.119\textwidth}
        \includegraphics[width=\linewidth]{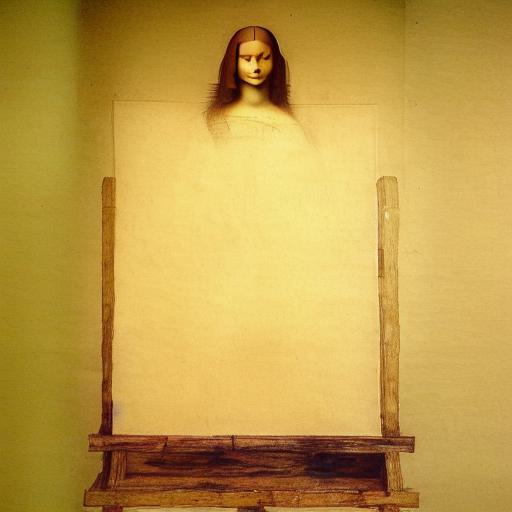}
    \end{minipage}
    \hfill
    \begin{minipage}[t]{0.119\textwidth}
        \includegraphics[width=\linewidth]{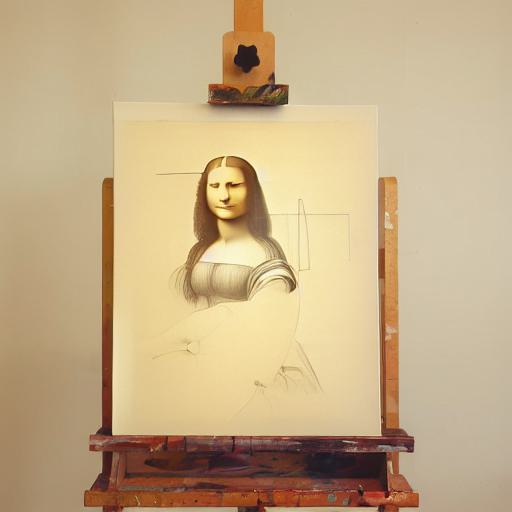}
    \end{minipage}
    \hfill
    \begin{minipage}[t]{0.119\textwidth}
        \includegraphics[width=\linewidth]{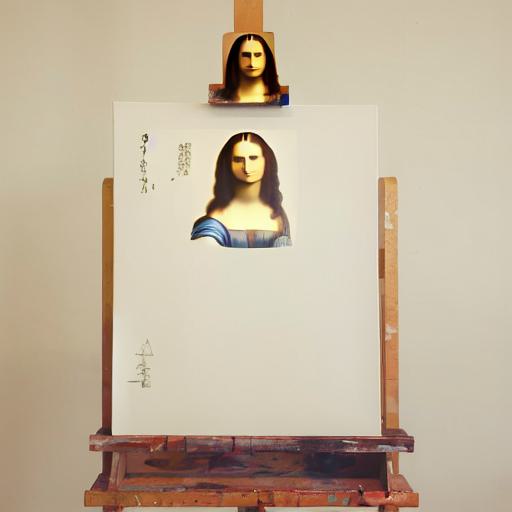}
    \end{minipage}
    \hfill
    \begin{minipage}[t]{0.119\textwidth}
        \includegraphics[width=\linewidth]{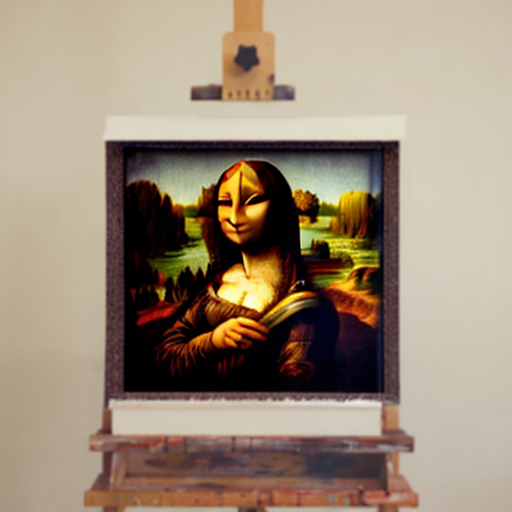}
    \end{minipage}
    \hfill
    \begin{minipage}[t]{0.119\textwidth}
        \includegraphics[width=\linewidth]{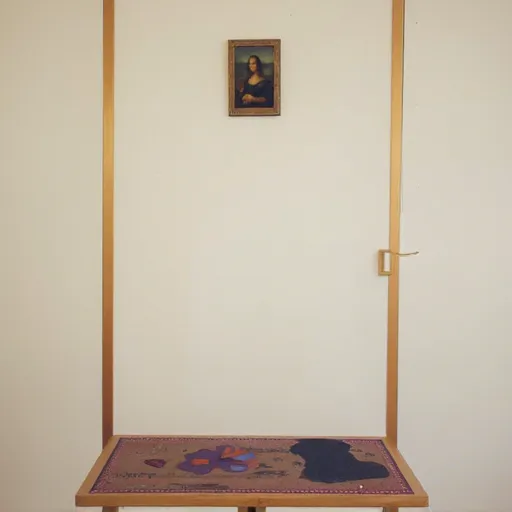}
    \end{minipage}
    \hfill
    \begin{minipage}[t]{0.119\textwidth}
        \includegraphics[width=\linewidth]{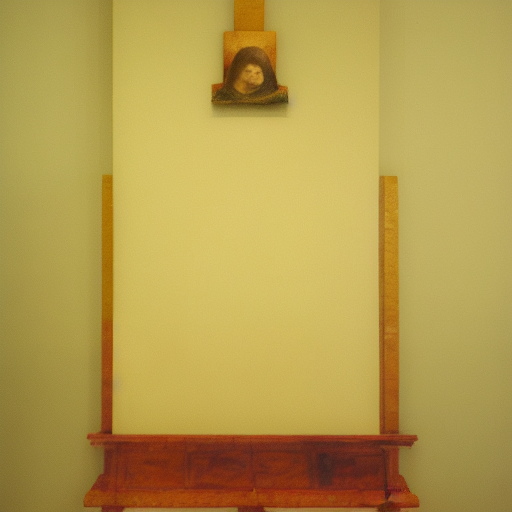}
    \end{minipage}
    \hfill
    \begin{minipage}[t]{0.119\textwidth}
        \includegraphics[width=\linewidth]{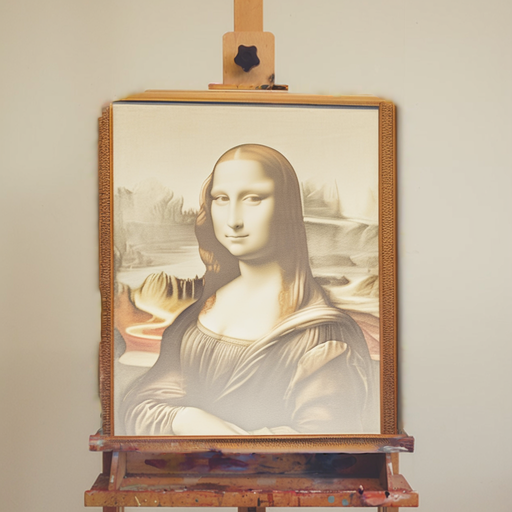}
    \end{minipage}
    \hfill

    % \par\vspace{0.3em}
    \scriptsize Draw Leonardo da Vinci's 'Mona Lisa' to the canvas

    \begin{minipage}[t]{0.119\textwidth}
        \includegraphics[width=\linewidth]{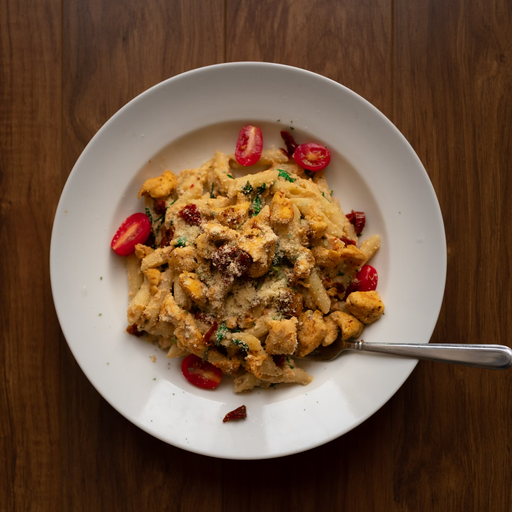}
    \end{minipage}
    \hfill
    \begin{minipage}[t]{0.119\textwidth}
        \includegraphics[width=\linewidth]{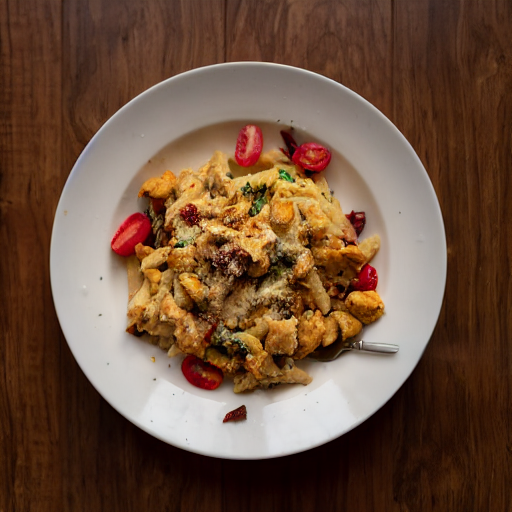}
    \end{minipage}
    \hfill
    \begin{minipage}[t]{0.119\textwidth}
        \includegraphics[width=\linewidth]{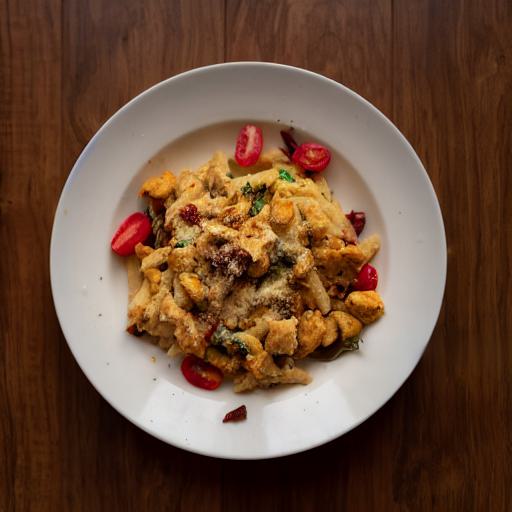}
    \end{minipage}
    \hfill
    \begin{minipage}[t]{0.119\textwidth}
        \includegraphics[width=\linewidth]{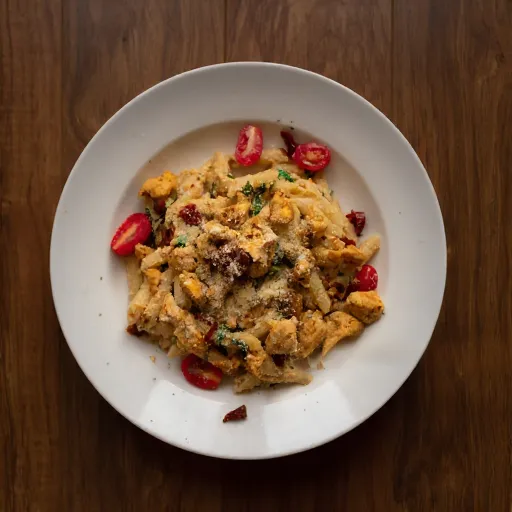}
    \end{minipage}
    \hfill
    \begin{minipage}[t]{0.119\textwidth}
        \includegraphics[width=\linewidth]{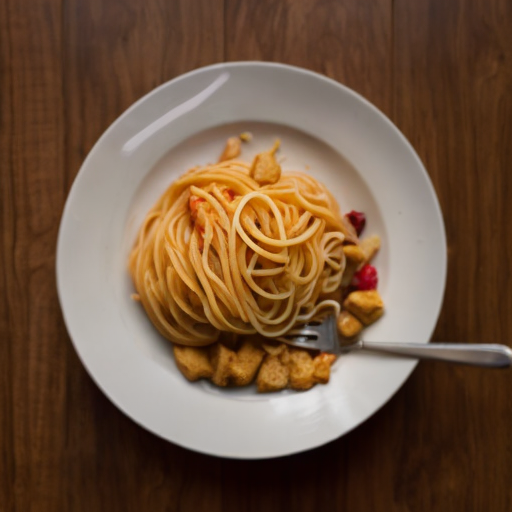}
    \end{minipage}
    \hfill
    \begin{minipage}[t]{0.119\textwidth}
        \includegraphics[width=\linewidth]{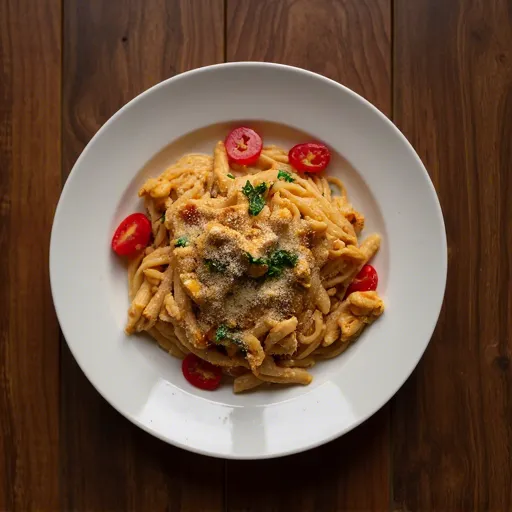}
    \end{minipage}
    \hfill
    \begin{minipage}[t]{0.119\textwidth}
        \includegraphics[width=\linewidth]{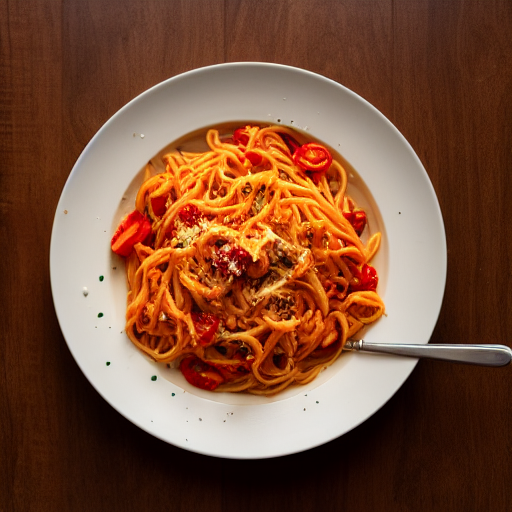}
    \end{minipage}
    \hfill
    \begin{minipage}[t]{0.119\textwidth}
        \includegraphics[width=\linewidth]{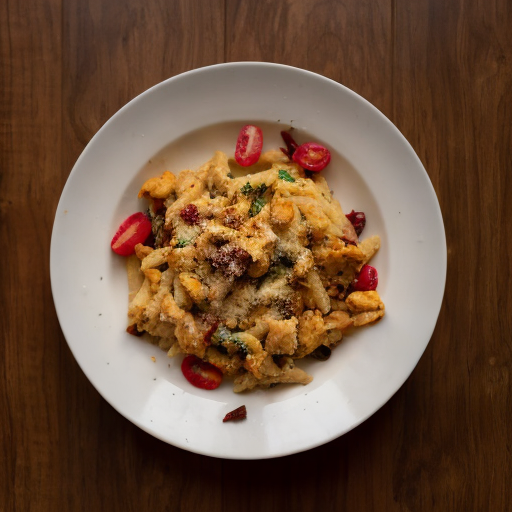}
    \end{minipage}

    % \par\vspace{0.3em}
    \scriptsize Remove the fork

    \begin{minipage}[t]{0.119\textwidth}
        \includegraphics[width=\linewidth]{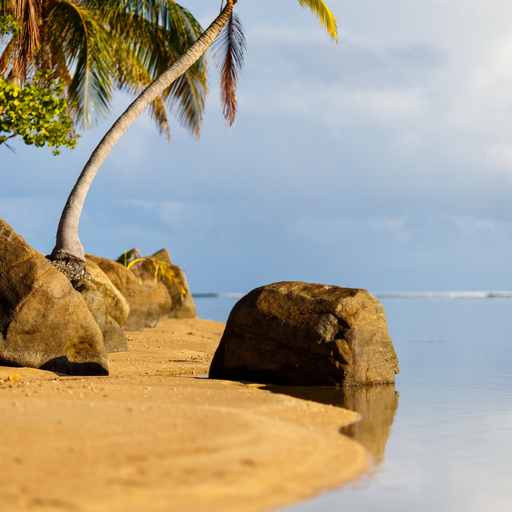}
    \end{minipage}
    \hfill
    \begin{minipage}[t]{0.119\textwidth}
        \includegraphics[width=\linewidth]{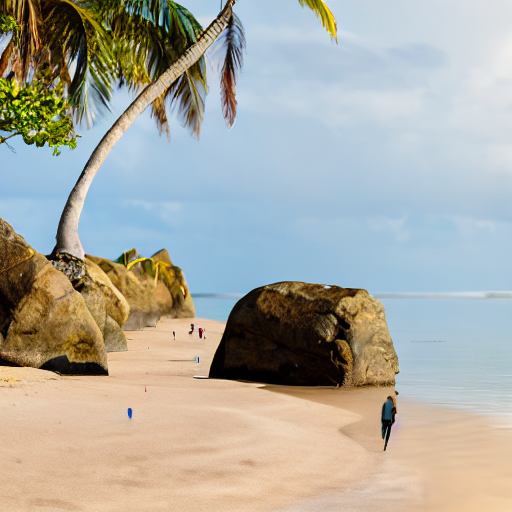}
    \end{minipage}
    \hfill
    \begin{minipage}[t]{0.119\textwidth}
        \includegraphics[width=\linewidth]{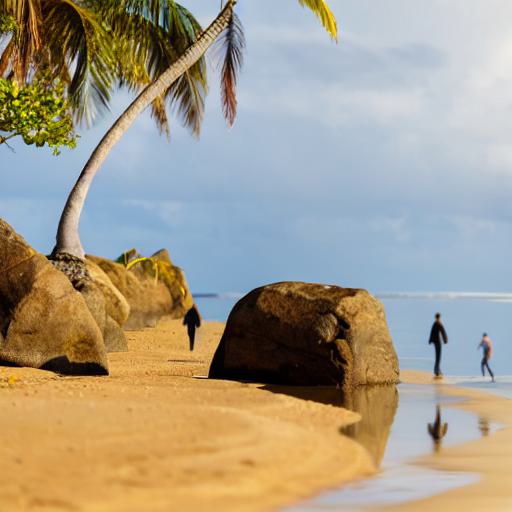}
    \end{minipage}
    \hfill
    \begin{minipage}[t]{0.119\textwidth}
        \includegraphics[width=\linewidth]{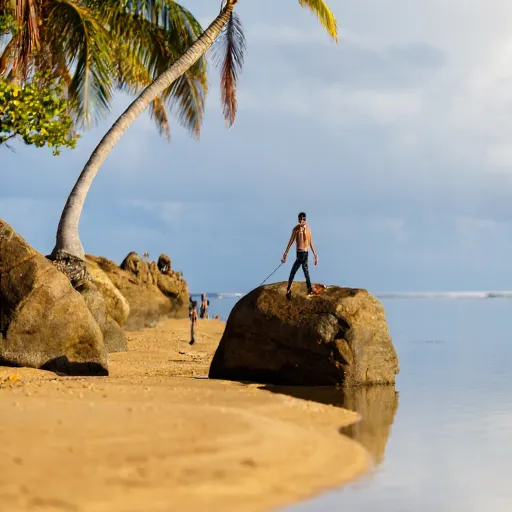}
    \end{minipage}
    \hfill
    \begin{minipage}[t]{0.119\textwidth}
        \includegraphics[width=\linewidth]{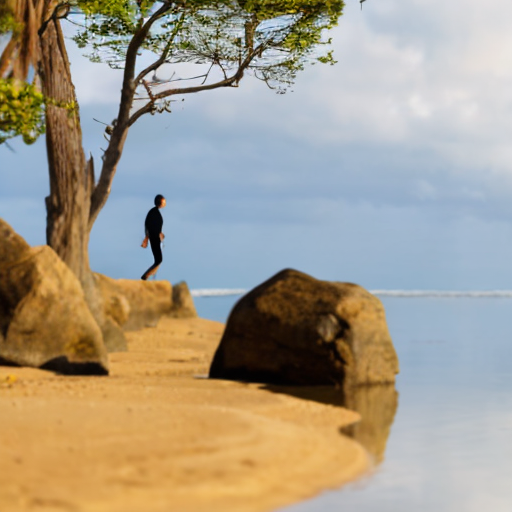}
    \end{minipage}
    \hfill
    \begin{minipage}[t]{0.119\textwidth}
        \includegraphics[width=\linewidth]{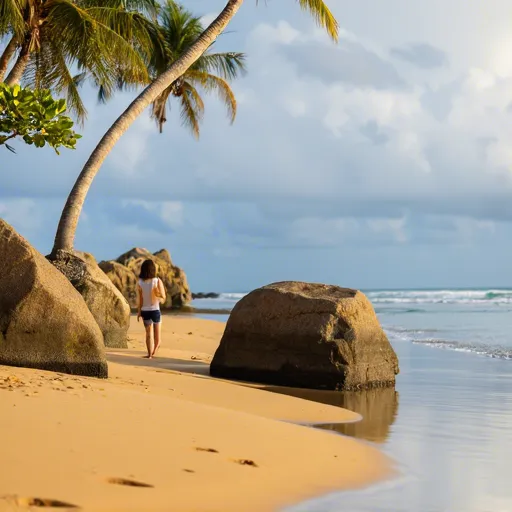}
    \end{minipage}
    \hfill
    \begin{minipage}[t]{0.119\textwidth}
        \includegraphics[width=\linewidth]{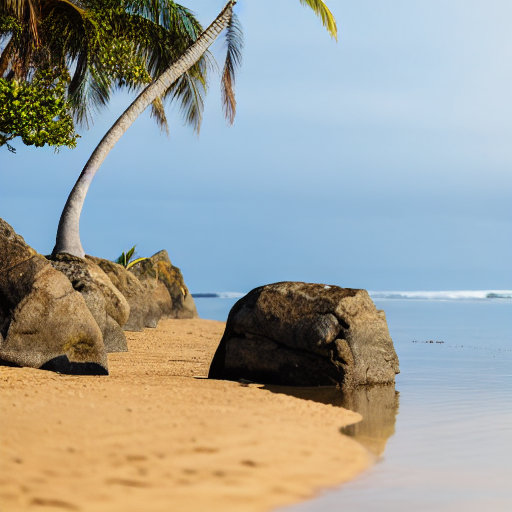}
    \end{minipage}
    \hfill
    \begin{minipage}[t]{0.119\textwidth}
        \includegraphics[width=\linewidth]{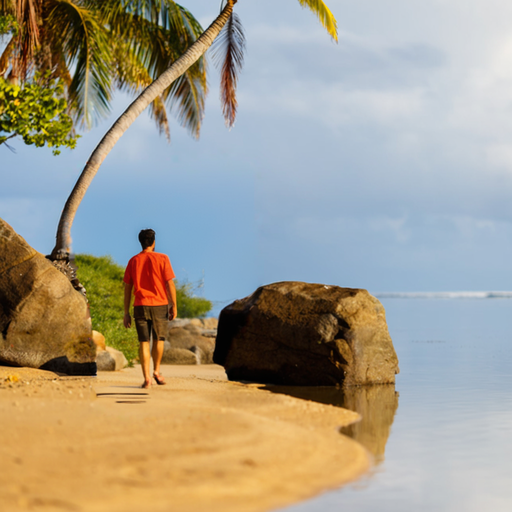}
    \end{minipage}

    % \par\vspace{0.3em}
    \scriptsize Add a pedestrian walking along the beach

    \begin{minipage}[t]{0.119\textwidth}
        \includegraphics[width=\linewidth]{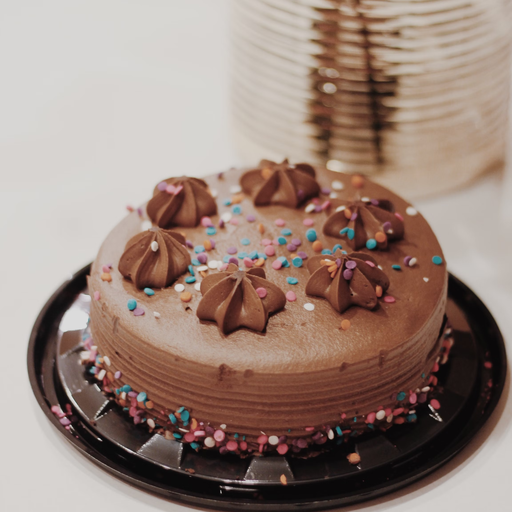}
    \end{minipage}
    \hfill
    \begin{minipage}[t]{0.119\textwidth}
        \includegraphics[width=\linewidth]{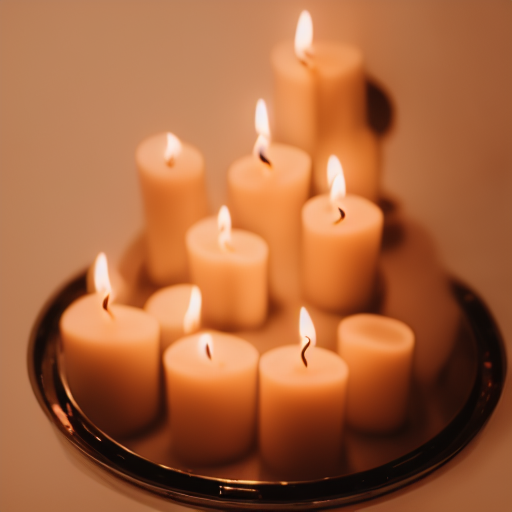}
    \end{minipage}
    \hfill
    \begin{minipage}[t]{0.119\textwidth}
        \includegraphics[width=\linewidth]{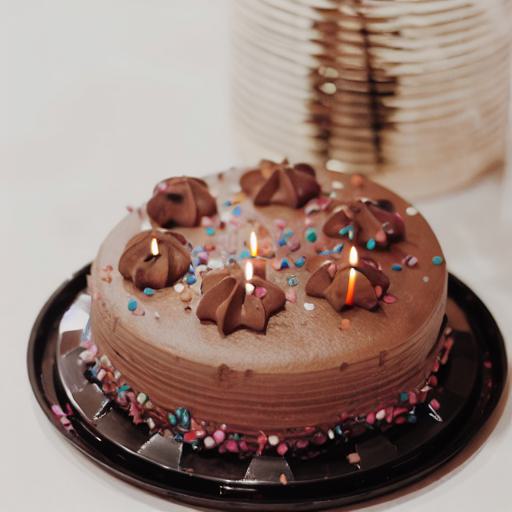}
    \end{minipage}
    \hfill
    \begin{minipage}[t]{0.119\textwidth}
        \includegraphics[width=\linewidth]{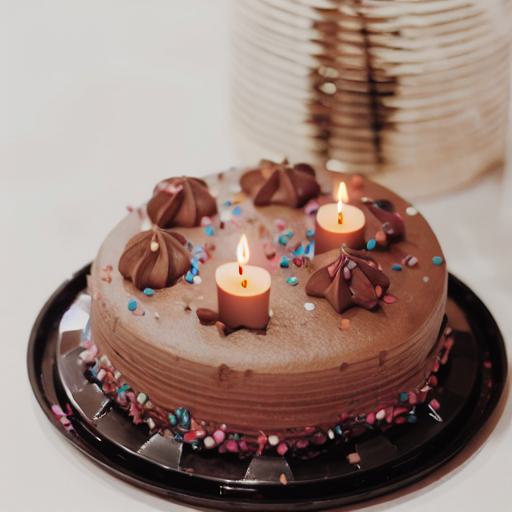}
    \end{minipage}
    \hfill
    \begin{minipage}[t]{0.119\textwidth}
        \includegraphics[width=\linewidth]{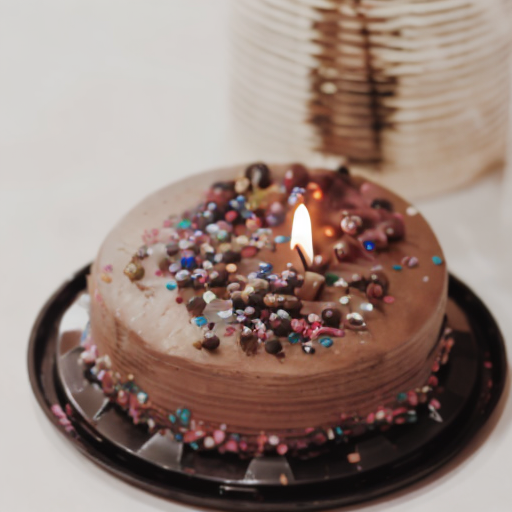}
    \end{minipage}
    \hfill
    \begin{minipage}[t]{0.119\textwidth}
        \includegraphics[width=\linewidth]{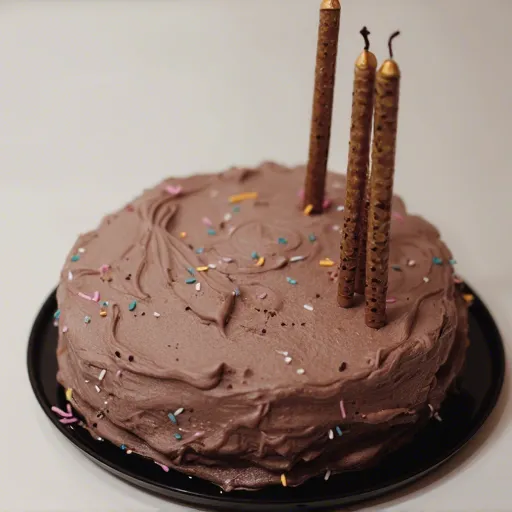}
    \end{minipage}
    \hfill
    \begin{minipage}[t]{0.119\textwidth}
        \includegraphics[width=\linewidth]{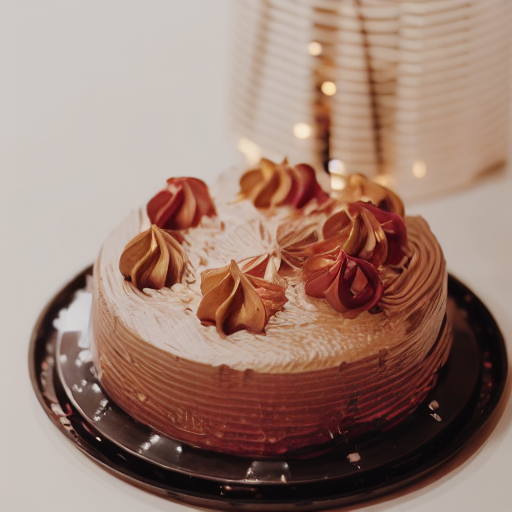}
    \end{minipage}
    \hfill
    \begin{minipage}[t]{0.119\textwidth}
        \includegraphics[width=\linewidth]{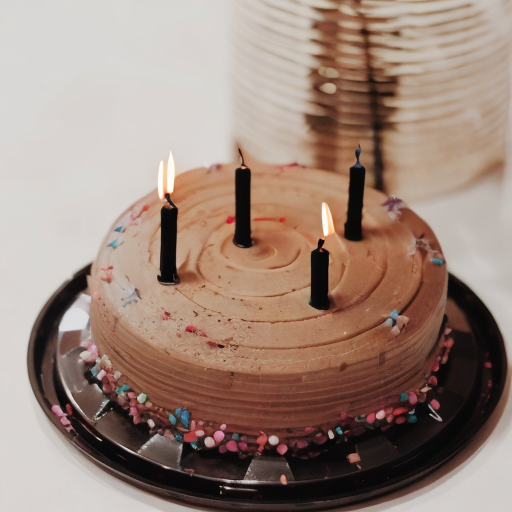}
    \end{minipage}

    % \par\vspace{0.3em}
    \scriptsize Replace the cookies with candles

    \begin{minipage}[t]{0.119\textwidth}
    \includegraphics[width=\linewidth]{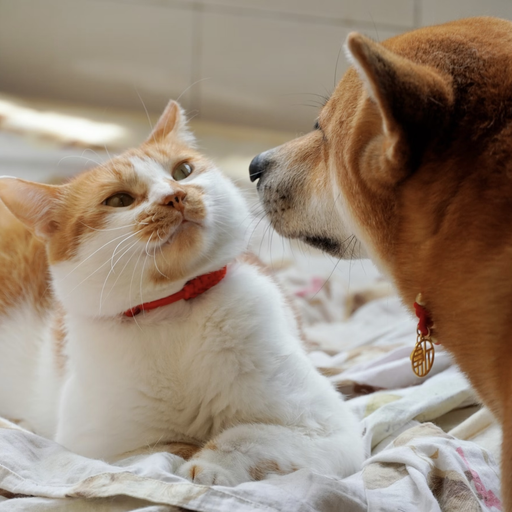}
    \end{minipage}
    \hfill
    \begin{minipage}[t]{0.119\textwidth}
        \includegraphics[width=\linewidth]{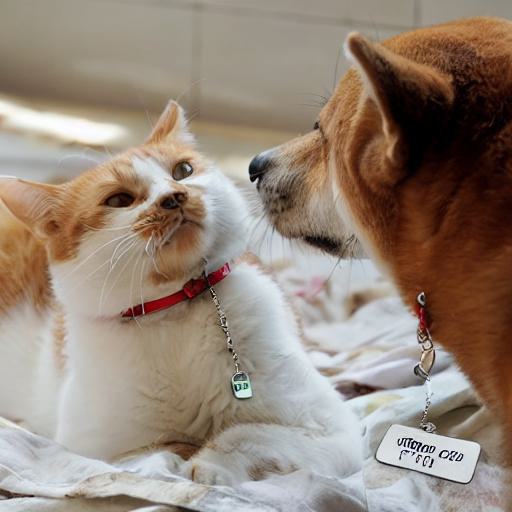}
    \end{minipage}
    \hfill
    \begin{minipage}[t]{0.119\textwidth}
        \includegraphics[width=\linewidth]{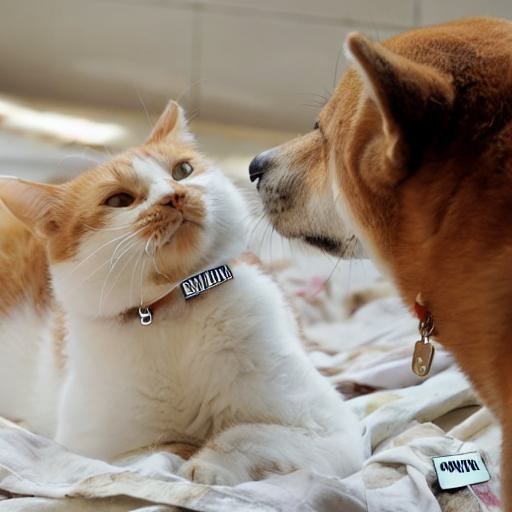}
    \end{minipage}
    \hfill
    \begin{minipage}[t]{0.119\textwidth}
        \includegraphics[width=\linewidth]{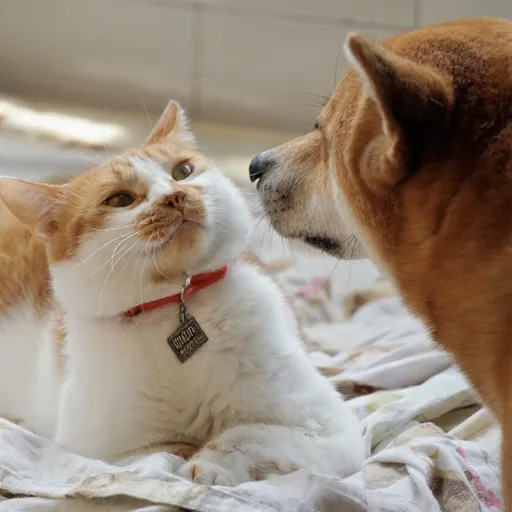}
    \end{minipage}
    \hfill
    \begin{minipage}[t]{0.119\textwidth}
        \includegraphics[width=\linewidth]{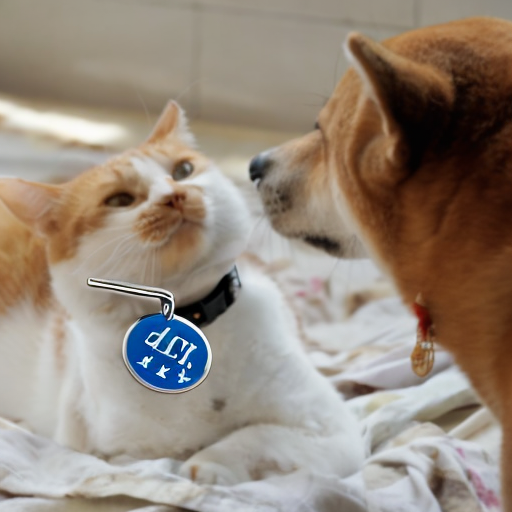}
    \end{minipage}
    \hfill
    \begin{minipage}[t]{0.119\textwidth}
        \includegraphics[width=\linewidth]{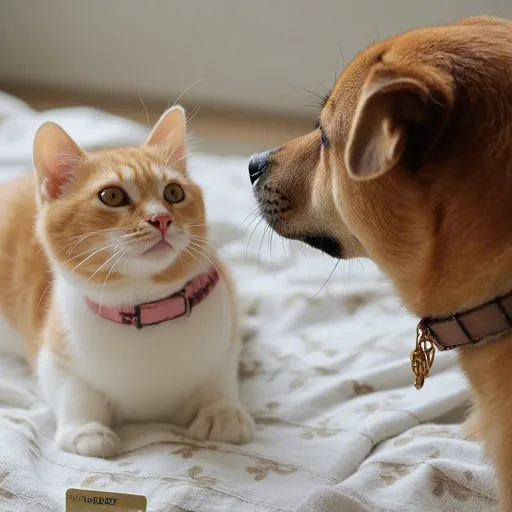}
    \end{minipage}
    \hfill
    \begin{minipage}[t]{0.119\textwidth}
        \includegraphics[width=\linewidth]{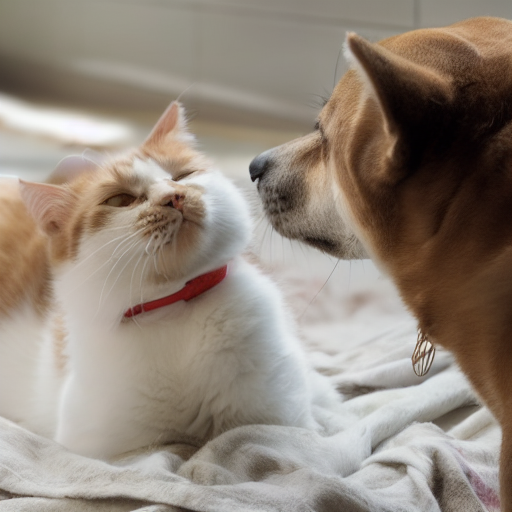}
    \end{minipage}
    \hfill
    \begin{minipage}[t]{0.119\textwidth}
        \includegraphics[width=\linewidth]{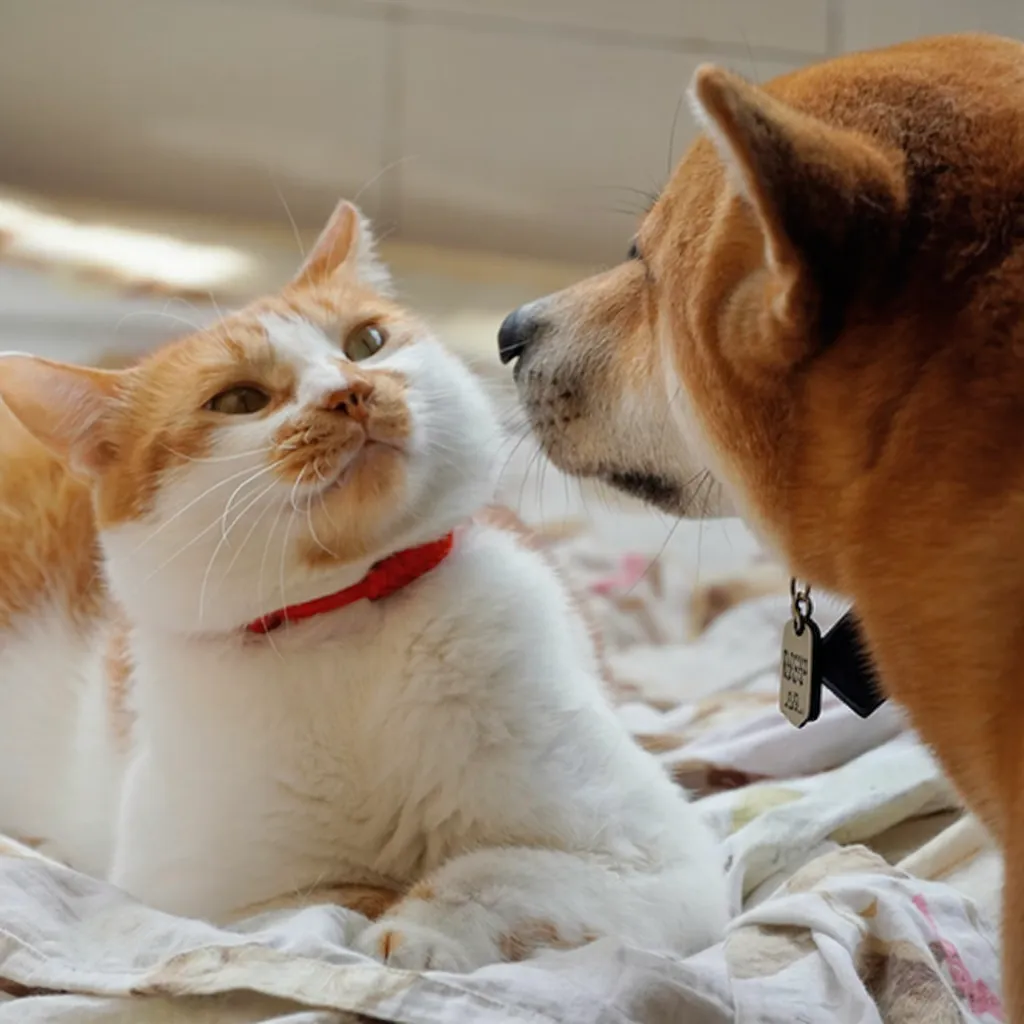}
    \end{minipage}

    % \par\vspace{0.3em}
    \scriptsize Change the pendant in the dog to a metal name tag

    % You would repeat the above 8-image block 3 more times for 4 groups in total

    \caption{\textbf{Comparison of editing results produced by different methods}. }
    \label{fig:comparison}
    \vspace{-2em}
\end{figure}

We apply our method to a variety of editing scenarios involving different objects and diverse editing operations, including addition, removal, and modification. The results shown in Figure~\ref{fig:comparison} and Figure~\ref{fig:variousOperation} demonstrate that our method can effectively handle a wide range of editing tasks across different contexts, producing high-quality edited images. Notably, the changes are localized within the learnable region, avoiding unintended modifications to the rest of the image. Furthermore, our method does not require paired editing datasets to perform instruction-driven image editing.
Due to the varying requirements of different editing operations and target objects, the learnable region extracts information from both the edit instruction and the image, and flexibly generates multi-scale editing regions to accommodate diverse editing scenarios, as shown in Figure~\ref{fig:multiScale}.

\subsection{Quantitative Evaluation}

To quantitatively assess our method, we evaluate its performance on two widely used instruction-based image editing benchmarks: Emu Edit~\cite{sheynin2024emu} and MagicBrush~\cite{NEURIPS2023_64008fa3}, both of which provide well-established evaluation tasks.
The MagicBrush benchmark assesses editing quality by comparing the generated images with ground truth images and their corresponding captions. Following the evaluation setup in MagicBrush~\cite{NEURIPS2023_64008fa3}, we use the L1 distance, L2 distance, CLIP image similarity, and DINO similarity as quantitative metrics.
The Emu Edit benchmark evaluates models based on how well the edited images align with the source image and the target textual description. Consistent with Emu Edit’s setup~\cite{sheynin2024emu}, we adopt several evaluation metrics: L1 distance, CLIP image similarity (CLIPimg), and DINO similarity to assess content preservation; CLIP text-image similarity (CLIPout), which measures the consistency between the edited image and the target caption; and CLIP text-image direction similarity (CLIPdir), which evaluates whether the editing reflects the semantic changes between the source and target captions.

The results on the Emu Edit Test~\cite{sheynin2024emu} are presented in Table~\ref{tab:EmuEdit}. We progressively increase the size of the training dataset from 1M to 5M samples. It is worth noting that this represents only a small fraction of the vast amount of available text-image pairs, and the dataset size can be easily scaled up further. As the training dataset grows, both editing performance metrics—CLIP direction similarity (CLIPdir) and CLIP text-image similarity (CLIPout)—and content preservation metrics—L1 distance, CLIP image similarity (CLIPimg), and DINO similarity—consistently improve. Our method achieves instructional image editing performance comparable to state-of-the-art approaches, despite being trained without any editing pairs dataset, while also demonstrating strong scalability.

\begin{table}[ht]
\vspace{-1em}
\centering
\caption{\textbf{Results on the Emu Edit Test}. We present benchmark results of models trained on varying scales of data and compare them with several existing methods. Best in \textbf{bold}, second in \textcolor{red}{red}. The results show that our method exhibits a consistent upward trend as the training data scale increases. Notably, our approach is trained using widely available text-image pairs rather than editing pairs, as discussed in the \hyperref[intro]{Introduction}, and the scale of such training data can be further expanded easily.}

\begin{tabular}{lccccc}
\toprule
\textbf{Method} & \textbf{CLIPdir$\uparrow$} & \textbf{CLIPout$\uparrow$} & \textbf{L1$\downarrow$} & \textbf{CLIPimg$\uparrow$} & \textbf{DINO$\uparrow$} \\
\midrule
InstructPix2Pix~\cite{Brooks_2023_CVPR} (450K) & 0.0782 & 0.2648 & 0.1217 & 0.8448 & 0.7643 \\
MagicBrush~\cite{NEURIPS2023_64008fa3} (450+20K)   & 0.0788 & 0.2749 & \textcolor{red}{0.0752} & \textbf{0.8979} & \textcolor{red}{0.8324} \\
UltraEdit~\cite{NEURIPS2024_05a30a0f} (3M)     & \textcolor{red}{0.1076} & \textcolor{red}{0.2832} & 0.0783 & 0.8436 & 0.7937 \\
LearnRegion~\cite{Lin_2024_CVPR} & 0.0671 & 0.2639 & 0.1113 & 0.8479 & 0.7437 \\
RF-Solver~\cite{wang2024taming} & 0.0834 & 0.2729 & 0.0767 & 0.8656 & 0.7962 \\
Plug-and-Play~\cite{tumanyan2023plug} & 0.0507 & 0.2342 & 0.1731 & 0.7964 & 0.7246 \\
\midrule
Ours (1M)     & 0.0792 & 0.2684 & 0.1215 & 0.8574 & 0.7856 \\
Ours (2M)     & 0.0884 & 0.2732 & 0.1008 & 0.8679 & 0.7989 \\
Ours (3M)     & 0.0979 & 0.2774 & 0.0843 & 0.8764 & 0.8104 \\
Ours (4M)     & 0.1026 & 0.2811 & 0.0779 & 0.8857 & 0.8217 \\
\rowcolor{gray!20}
Ours (5M)     & \textbf{0.1088} & \textbf{0.2842} & \textbf{0.0723} & \textcolor{red}{0.8913} & \textbf{0.8337} \\
\bottomrule
\end{tabular}
\label{tab:EmuEdit}
\vspace{-1em}

\end{table}

The results on the MagicBrush test set, shown in Table~\ref{tab:MagicBrush}, demonstrate that our approach effectively handles both single-turn and multi-turn editing tasks. It achieves performance comparable to methods trained on editing pairs, and shows particular strength in following semantic instructions accurately.

\begin{table}[ht]
\vspace{-1em}

\centering
\caption{\textbf{Results on the MagicBrush test set}. We report performance under both single-turn and multi-turn settings. Best results are shown in \textbf{bold}, and second-best in \textcolor{red}{red}. Our method achieves performance comparable to state-of-the-art approaches in both settings without editing pairs dataset.}

\begin{tabular}{llcccc}
\toprule
\textbf{Settings} & \textbf{Methods} & \textbf{L1$\downarrow$} & \textbf{L2$\downarrow$} & \textbf{CLIP-I$\uparrow$} & \textbf{DINO$\uparrow$} \\
\midrule
\multirow{5}{*}{\shortstack[l]{Single-turn}} 
& InstructPix2Pix~\cite{Brooks_2023_CVPR} (450K) & 0.1137 & 0.0368 & 0.8521 & 0.7429 \\
& MagicBrush~\cite{NEURIPS2023_64008fa3} (450+20K) & \textbf{0.0618} & \textcolor{red}{0.0205} & 0.9324 & \textbf{0.8979} \\
& UltraEdit~\cite{NEURIPS2024_05a30a0f} (3M) & 0.0689 & \textbf{0.0201} & \textcolor{red}{0.8986} & 0.8477 \\
& RF-Solver~\cite{wang2024taming} & 0.0787 & 0.312 & 0.8768 & 0.8214 \\
\rowcolor{gray!20}
& Ours (5M) & \textcolor{red}{0.0657} & 0.0212 & \textbf{0.9383} & \textcolor{red}{0.8589} \\
\midrule
\multirow{5}{*}{\shortstack[l]{Multi-turn}} 
& InstructPix2Pix~\cite{Brooks_2023_CVPR} (450K) & 0.1339 & 0.0455 & 0.8294 & 0.7002 \\
& MagicBrush~\cite{NEURIPS2023_64008fa3} (450+20K) & \textcolor{red}{0.0957} & 0.0353 & \textcolor{red}{0.8922} & \textbf{0.8271} \\
& UltraEdit~\cite{NEURIPS2024_05a30a0f} (3M) & \textbf{0.0912} & \textbf{0.0323} & 0.8594 & 0.7842 \\
& RF-Solver~\cite{wang2024taming} & 0.1176 & 0.0423 & 0.8304 & 0.7079 \\
\rowcolor{gray!20}
& Ours (5M) & 0.0972 & \textcolor{red}{0.0332} & \textbf{0.8985} & \textcolor{red}{0.7962} \\
\bottomrule
\end{tabular}
\label{tab:MagicBrush}
\vspace{-1em}

\end{table}

\subsection{Compatibility with Various Generative Models} \label{compatibility}

\begin{figure}[H]
    \vspace{-1em}

    \centering

    \begin{minipage}[t]{0.49\textwidth}
        \centering
        \begin{minipage}[t]{0.24\textwidth}
            \includegraphics[width=\linewidth]{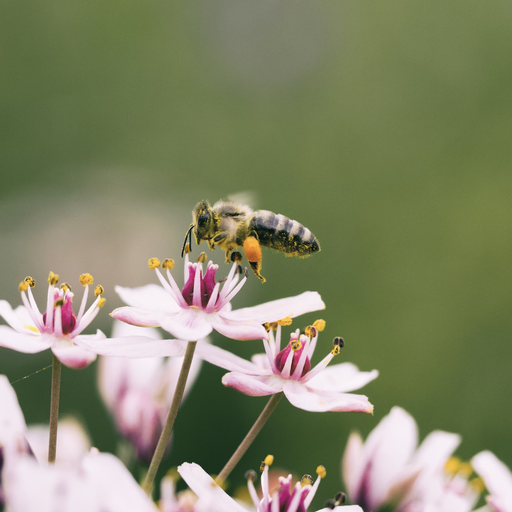}
        \end{minipage}\hfill
        \begin{minipage}[t]{0.24\textwidth}
            \includegraphics[width=\linewidth]{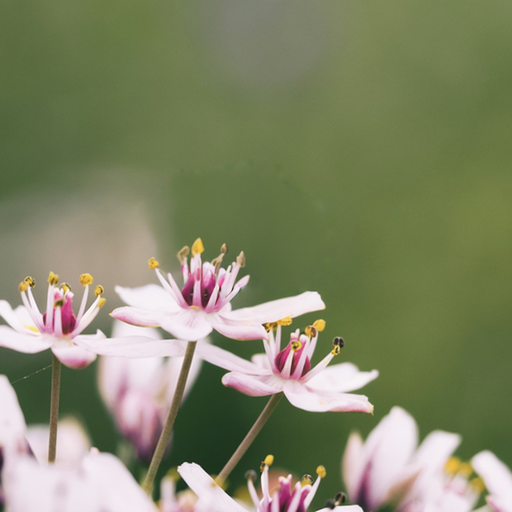}
        \end{minipage}\hfill
        \begin{minipage}[t]{0.24\textwidth}
            \includegraphics[width=\linewidth]{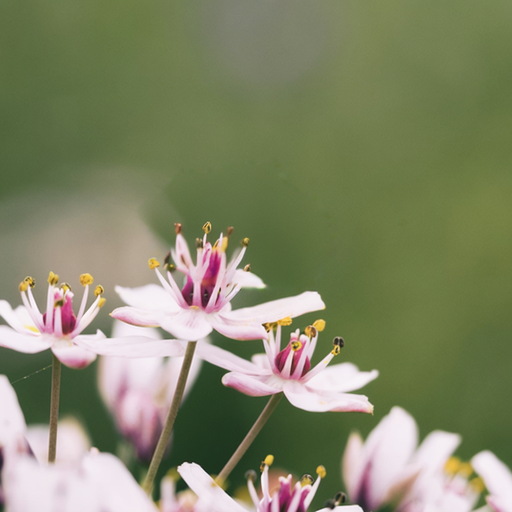}
        \end{minipage}\hfill
        \begin{minipage}[t]{0.24\textwidth}
            \includegraphics[width=\linewidth]{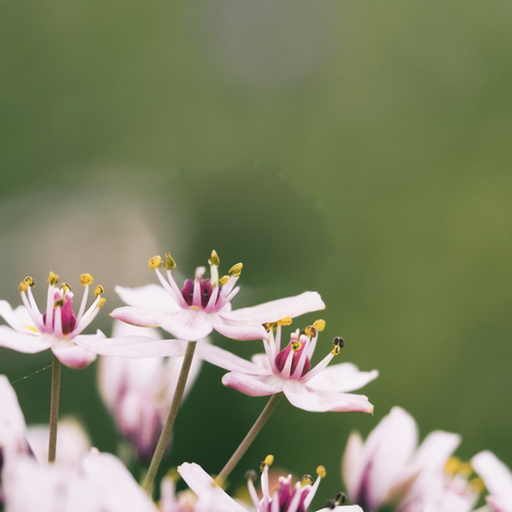}
        \end{minipage}
        \scriptsize\\ Remove the bee
    \end{minipage}
    % \hfill
    \begin{minipage}[t]{0.49\textwidth}
        \centering
        \begin{minipage}[t]{0.24\textwidth}
            \includegraphics[width=\linewidth]{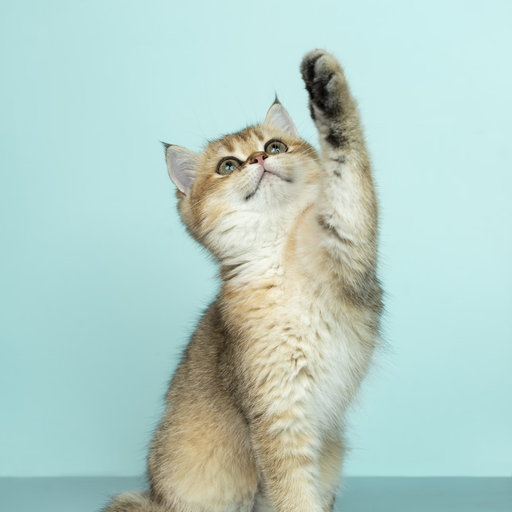}
        \end{minipage}\hfill
        \begin{minipage}[t]{0.24\textwidth}
            \includegraphics[width=\linewidth]{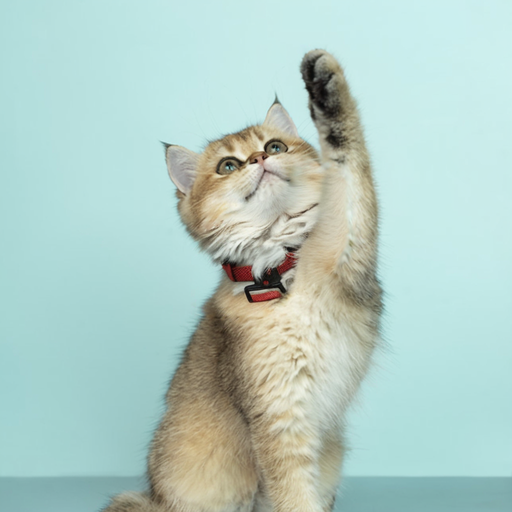}
        \end{minipage}\hfill
        \begin{minipage}[t]{0.24\textwidth}
            \includegraphics[width=\linewidth]{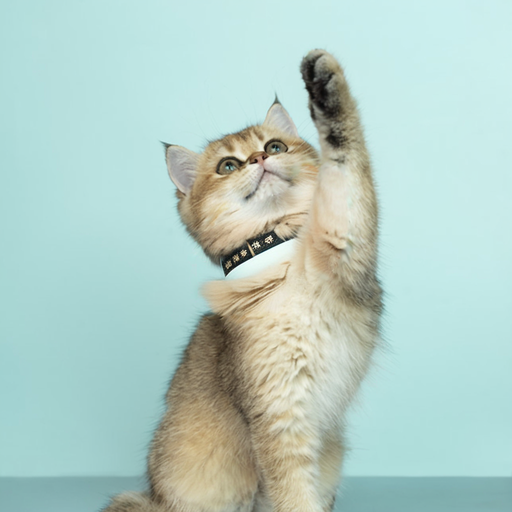}
        \end{minipage}\hfill
        \begin{minipage}[t]{0.24\textwidth}
            \includegraphics[width=\linewidth]{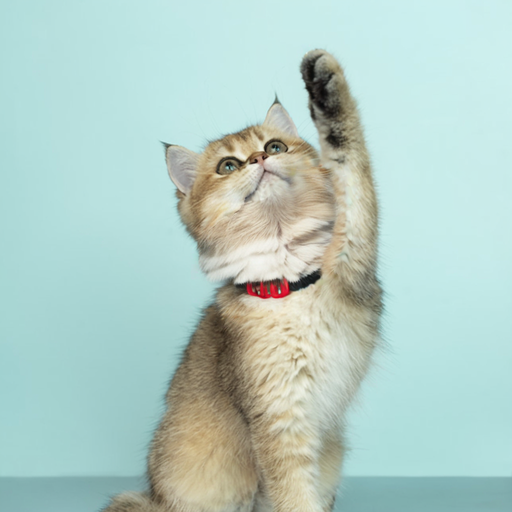}
        \end{minipage}
        \scriptsize\\ Add the cat a collar
    \end{minipage}

    \caption{\textbf{Editing results produced by our method using different generative models}. For each example, from left to right: original image, FLUX~\cite{flux2024}, VAR~\cite{tian2024visual}, and MaskGIT~\cite{chang2022maskgit}.}

    \label{fig:compatibility}
    \vspace{-1em}
\end{figure}

As illustrated in \hyperref[ImplementDetails]{Implementation Details}, our method is compatible with various generative models, as shown in Figure~\ref{fig:compatibility}, including FLUX~\cite{flux2024}, VAR~\cite{tian2024visual}, and MaskGIT~\cite{chang2022maskgit}, demonstrating the broad applicability of our approach.

\subsection{Ablation Study}

\begin{figure}[htbp]
    \vspace{-1em}

    \centering

    % --- Row of 7 Images ---
    % Image 1: Original
    \begin{minipage}[t]{0.14\textwidth}
        \centering
        \includegraphics[width=\linewidth]{fig/23.png}
        % \vspace{0.1em} % Small space between image and caption
        % \scriptsize Original
    \end{minipage}%
    \hspace{0.005\textwidth}%
    % Image 2: wo_semantic_mask
    \begin{minipage}[t]{0.14\textwidth}
        \centering
        \includegraphics[width=\linewidth]{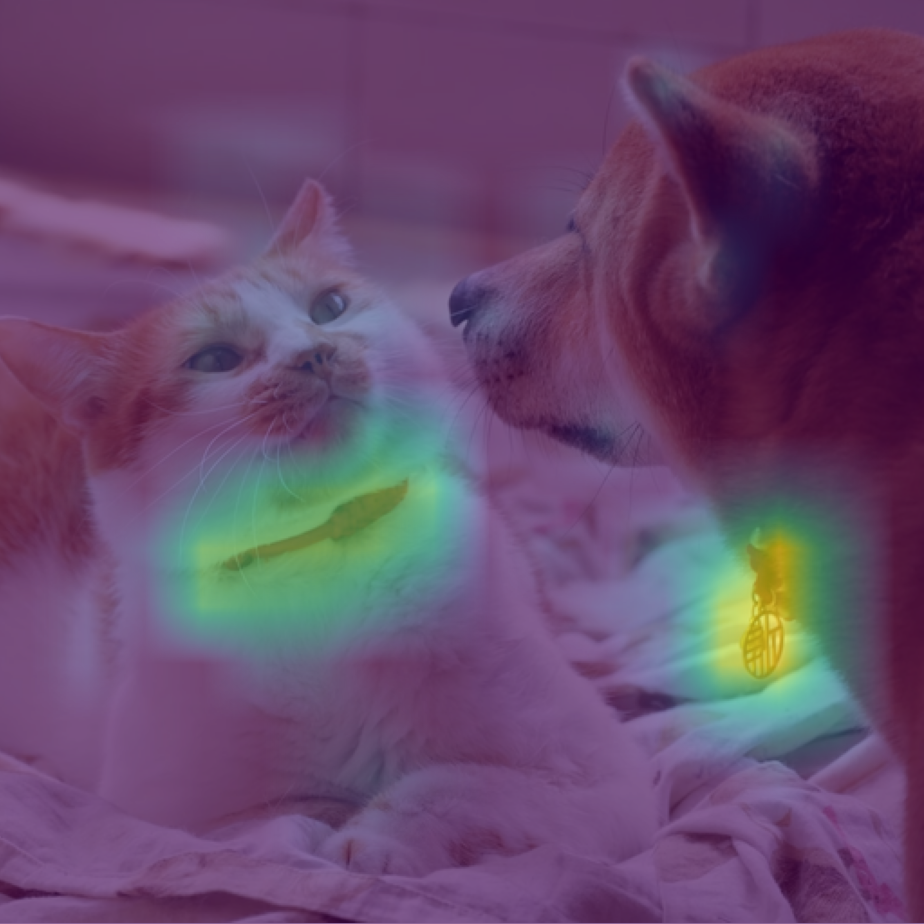}
        % This minipage only contains an image
    \end{minipage}%
    % \hspace{0.005\textwidth}%
    % Image 3: wo_semantic
    \begin{minipage}[t]{0.14\textwidth}
        \centering
        \includegraphics[width=\linewidth]{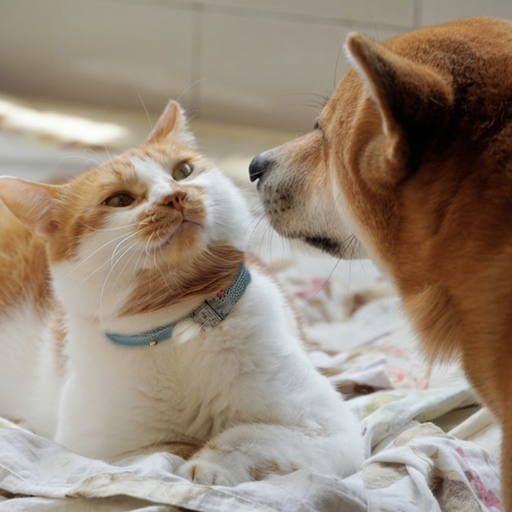}
    \end{minipage}%
    \hspace{0.005\textwidth}%
    % Image 4: wo_CLIP_mask
    \begin{minipage}[t]{0.14\textwidth}
        \centering
        \includegraphics[width=\linewidth]{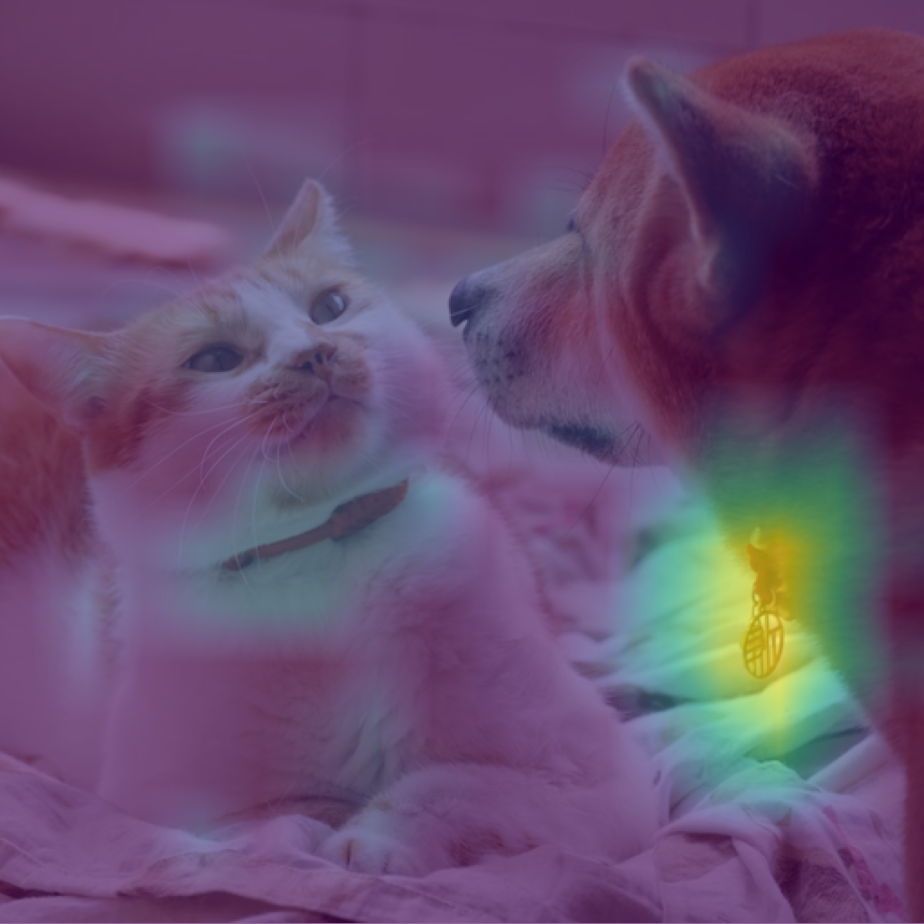}
    \end{minipage}%
    % \hspace{0.005\textwidth}%
    % Image 5: wo_CLIP
    \begin{minipage}[t]{0.14\textwidth}
        \centering
        \includegraphics[width=\linewidth]{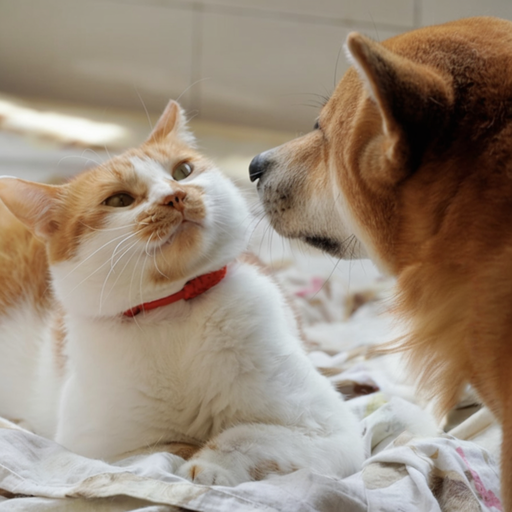}
    \end{minipage}%
    \hspace{0.005\textwidth}%
    % Image 6: full_ours_mask
    \begin{minipage}[t]{0.14\textwidth}
        \centering
        \includegraphics[width=\linewidth]{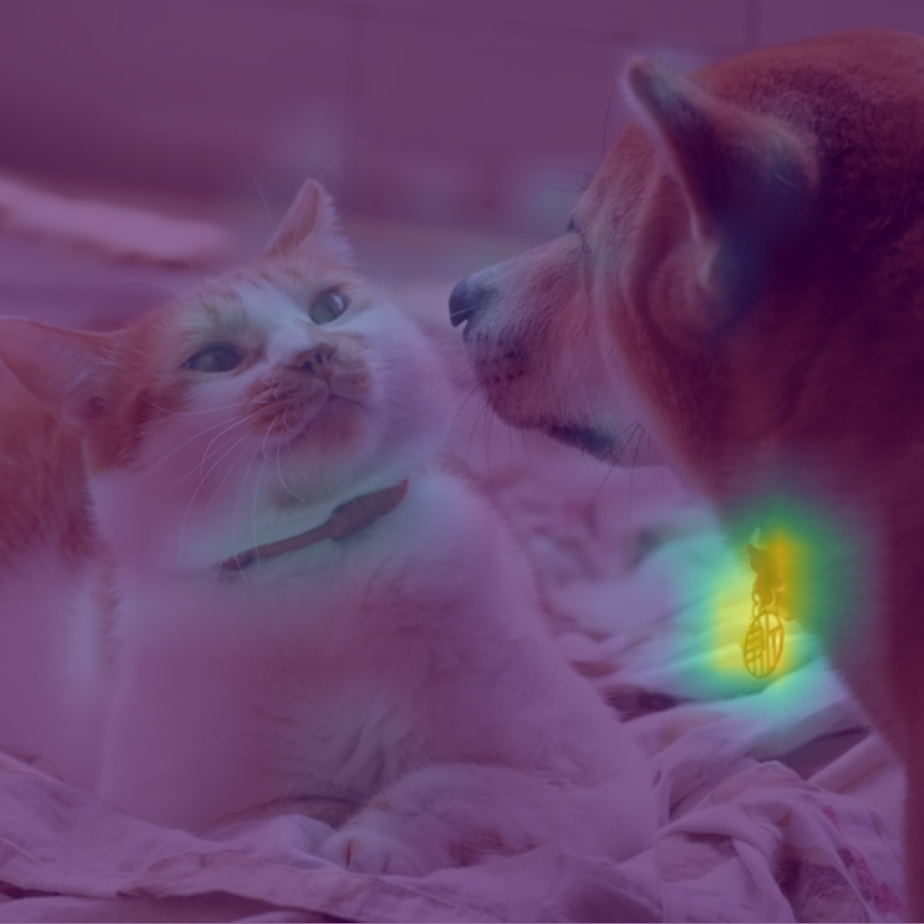}
    \end{minipage}%
    % \hspace{0.005\textwidth}%
    % Image 7: full_ours
    \begin{minipage}[t]{0.14\textwidth}
        \centering
        \includegraphics[width=\linewidth]{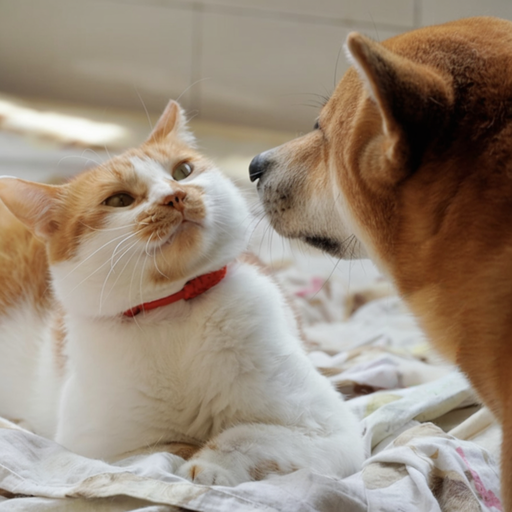}
    \end{minipage} % No \hspace after the last minipage in the row

    % --- Group labels under image pairs ---
    % \par\vspace{0.3em}
    % \hspace{0.125\textwidth}% % skip "Original" image space (0.12\textwidth image + part of first space)
    \begin{minipage}[t]{0.14\textwidth}
        \centering
        \scriptsize Original Image
    \end{minipage}%
    \begin{minipage}[t]{0.28\textwidth}
        \centering
        \scriptsize w/o $L_\text{semAlign}$
    \end{minipage}%
    \hspace{0.005\textwidth}%
    \begin{minipage}[t]{0.28\textwidth}
        \centering
        \scriptsize w/o $L_\text{CLIP}$
    \end{minipage}%
    \hspace{0.005\textwidth}%
    \begin{minipage}[t]{0.28\textwidth}
        \centering
        \scriptsize Ours
    \end{minipage} % No \hspace after the last minipage in the row

    \caption{\textbf{Influence of each loss component.} The editing instruction is: "Remove the pendant from the dog." From left to right: original image, result without $L_\text{semAlign}$, result without $L_\text{CLIP}$, and result using all loss terms. Each result includes both the learnable region and the edited image.}

    \label{fig:ablation}
    \vspace{-1em}

\end{figure}

To evaluate the influence of each loss component: $L_{\text{semAlign}}$~\eqref{eq:loss_semAlign} and $L_{\text{CLIP}}$~\eqref{eq:loss_CLIP}, we conduct an ablation study, as shown in Figure~\ref{fig:ablation} and Table~\ref{tab:ablation}. The results demonstrate that without $L_{\text{semAlign}}$, the learnable region fails to accurately localize the correct area, especially when multiple potential target regions (e.g., due to similar objects) are present in the image. On the other hand, removing $L_{\text{CLIP}}$ results in an excessively large and sparse learnable region. In contrast, our method with both loss terms produces accurate and appropriately scaled learnable regions for editing.

\section{Conclusion}
% In this work, we present a novel paradigm for instruction-driven image editing that circumvents the limitations of current methods reliant on large-scale editing pair datasets or training-free approaches with restricted capabilities. 
% Extensive experiments conducted across various benchmarks demonstrate that our method achieves performance comparable to state-of-the-art approaches, while requiring no retraining or fine-tuning of the generative model. Furthermore, our approach exhibits high adaptability to different types of generative backbones, indicating strong generalization and scalability.

We propose a novel instruction-driven image editing framework that eliminates the reliance on scarce editing-pair datasets and overcomes the limitations of dataset-free methods. By leveraging abundant and diverse text-image pairs, our approach enables precise, high-fidelity edits through multi-scale learnable regions, guided by vision-language alignment. Without retraining or fine-tuning the generative model, our method achieves state-of-the-art performance across benchmarks and generalizes well to various generative backbones. This work highlights a scalable and data-efficient alternative for instruction-based editing, paving the way for future research that fully harnesses existing multimodal resources.

{
    \small
    \bibliographystyle{ieeenat_fullname}
    \bibliography{main}
}

% {
% \small

% [1] Alexander, J.A.\ \& Mozer, M.C.\ (1995) Template-based algorithms for
% connectionist rule extraction. In G.\ Tesauro, D.S.\ Touretzky and T.K.\ Leen
% (eds.), {\it Advances in Neural Information Processing Systems 7},
% pp.\ 609--616. Cambridge, MA: MIT Press.

% [2] Bower, J.M.\ \& Beeman, D.\ (1995) {\it The Book of GENESIS: Exploring
%   Realistic Neural Models with the GEneral NEural SImulation System.}  New York:
% TELOS/Springer--Verlag.

% [3] Hasselmo, M.E., Schnell, E.\ \& Barkai, E.\ (1995) Dynamics of learning and
% recall at excitatory recurrent synapses and cholinergic modulation in rat
% hippocampal region CA3. {\it Journal of Neuroscience} {\bf 15}(7):5249-5262.
% }

%%%%%%%%%%%%%%%%%%%%%%%%%%%%%%%%%%%%%%%%%%%%%%%%%%%%%%%%%%%%
\newpage

\appendix

\section{Text-to-image Generative Model Conditioned on Region} \label{Appendix_A}

Different generative models are typically built upon distinct conditional control strategies, resulting in varied mathematical formulations. Here, we present the training procedure and formulation of editing-region-conditioned text-to-image generation using Stable Diffusion. For other generative models, including FLUX~\cite{flux2024}, VAR~\cite{tian2024visual}, and MaskGIT~\cite{chang2022maskgit}, please refer to their respective works.

\subsection{Preliminaries: Stable Diffusion Components}
The core components of Stable Diffusion are:

\begin{itemize} [leftmargin=10pt]
    \item \textbf{Image Encoder $\mathcal{E}$}: A neural network that maps an image $X \in \mathbb{R}^{H \times W \times 3}$ from pixel space to a lower-dimensional latent representation $z_0 = \mathcal{E}(X) \in \mathbb{R}^{h \times w \times c_z}$, where $f = H/h = W/w$ is the downsampling factor and $c_z$ is the number of latent channels.
    \item \textbf{Image Decoder $\mathcal{D}$}: A neural network that maps a latent representation $z \in \mathbb{R}^{h \times w \times c_z}$ back to an image $\hat{X} = \mathcal{D}(z) \in \mathbb{R}^{H \times W \times 3}$.
    \item \textbf{Text Encoder $\tau_{\text{text}}$}: Typically a pre-trained CLIP text encoder, which maps a text prompt $t \in \mathbb{S}$ (where $\mathbb{S}$ is the space of text strings) to a conditioning embedding $c_{\text{text}} = \tau_{\text{text}}(t) \in \mathbb{R}^{L \times d_c}$, where $L$ is the token sequence length and $d_c$ is the embedding dimension.
    \item \textbf{Denoising U-Net $\epsilon_\theta$}: A time-conditional U-Net architecture with parameters $\theta$. It operates in the latent space and is trained to predict the noise $\epsilon$ added to a noisy latent $z_t$ at diffusion timestep $t$. Its input includes $z_t$, $t$, and conditioning information $c$. For inpainting, $c$ can include $c_{\text{text}}$ and spatial conditioning derived from the mask and known image regions.
    \item \textbf{Forward Diffusion Process}: This process gradually adds Gaussian noise to a latent $z_0$ over $T$ timesteps according to a variance schedule $\beta_1, ..., \beta_T$. A noisy latent $z_t$ at timestep $t$ is given by:
    \begin{equation}
        q(z_t | z_0) = \mathcal{N}(z_t; \sqrt{\bar{\alpha}_t} z_0, (1-\bar{\alpha}_t)I)
        \label{eq:forward_process}
    \end{equation}
    where $\alpha_t = 1 - \beta_t$ and $\bar{\alpha}_t = \prod_{s=1}^{t} \alpha_s$. This allows direct sampling: $z_t = \sqrt{\bar{\alpha}_t} z_0 + \sqrt{1-\bar{\alpha}_t} \epsilon$, for $\epsilon \sim \mathcal{N}(0, I)$.
\end{itemize}

\subsection{Training a Region-Conditioned Stable Diffusion Model}
The objective is to train the U-Net $\epsilon_\theta$ to denoise $z_t$ while respecting the unmasked regions and the text prompt.

Given a dataset of images $X$, text prompts $t$, and masks $M_{\text{train}}$:
\begin{enumerate}[leftmargin=10pt]
    \item The original image $X$ is encoded to $z_0 = \mathcal{E}(X)$.
    \item The training mask $M_{\text{train}}$ is downsampled to the latent space resolution, $m_{\text{train}} \in \{0, 1\}^{h \times w}$. We define $m_{\text{train}}=1$ for regions to be inpainted and $m_{\text{train}}=0$ for regions to be preserved.
    \item The conditioning input for the U-Net is prepared. A common strategy for LDM inpainting \cite{rombach2022high} is to concatenate the noisy latent $z_t$ with the mask $m_{\text{train}}$ and the latent representation of the known (unmasked) regions $(1-m_{\text{train}}) \odot z_0$. Let $z_0^{\text{masked\_context}} = (1-m_{\text{train}}) \odot z_0$. The U-Net input at step $t$ becomes $z_t^{\text{inp}} = \text{concat}(z_t, m_{\text{train}}, z_0^{\text{masked\_context}})$.
    \item The text prompt $t$ is encoded: $c_{\text{text}} = \tau_{\text{text}}(t)$.
    \item A timestep $t \sim \mathcal{U}(\{1, ..., T\})$ and noise $\epsilon \sim \mathcal{N}(0, I)$ are sampled.
    \item The noisy latent is formed: $z_t = \sqrt{\bar{\alpha}_t} z_0 + \sqrt{1-\bar{\alpha}_t} \epsilon$.
    \item The U-Net predicts the noise: $\epsilon_{\text{pred}} = \epsilon_\theta(z_t^{\text{inp}}, t, c_{\text{text}})$.
\end{enumerate}
The model is trained by minimizing the loss:
\begin{align}
\mathcal{L}_{\text{LDM-inpainting}} = 
\mathbb{E}_{X, t, M_{\text{train}}, \epsilon, t} \big[ \big\| \epsilon - \epsilon_\theta(&\text{concat}(\sqrt{\bar{\alpha}_t}\mathcal{E}(X) \notag \\
&+ \sqrt{1 - \bar{\alpha}_t}\epsilon, \; m_{\text{train}}, \; (1 - m_{\text{train}}) \odot \mathcal{E}(X)), \; t, \; \tau_{\text{text}}(t)) \big\|_2^2 \big]
\label{eq:training_loss}
\end{align}

% \begin{equation}
%     \mathcal{L}_{\text{LDM-inpainting}} = \mathbb{E}_{X, t, M_{\text{train}}, \epsilon, t} \left[ || \epsilon - \epsilon_\theta(\text{concat}(\sqrt{\bar{\alpha}_t}\mathcal{E}(X) + \sqrt{1-\bar{\alpha}_t}\epsilon, m_{\text{train}}, (1-m_{\text{train}})\odot\mathcal{E}(X)), t, \tau_{\text{text}}(t)) ||_2^2 \right]
%     \label{eq:training_loss}
% \end{equation}
This trains $\epsilon_\theta$ to be aware of the mask and the content of the unmasked regions.

\subsection{Inference Phase for Instruction-Driven Image Editing}
Given the source image $X \equiv X_{\text{src}}$, the target text description $t_e$, and the predicted editing region mask $M_{\text{region}}$ (where $M_{\text{region}}(i,j)=1$ indicates the edit region):

\begin{enumerate}[leftmargin=10pt]
    \item \textbf{Initialization}:
    \begin{itemize}
        \item Encode the source image into its initial latent representation:
        \begin{equation}
            z_0^{\text{src}} = \mathcal{E}(X_{\text{src}})
        \end{equation}
        \item Encode the target text description into a conditioning embedding:
        \begin{equation}
            c_e = \tau_{\text{text}}(t_e)
        \end{equation}
        \item Downsample the edit region mask $M_{\text{region}}$ to the latent space resolution to obtain $m \in \{0, 1\}^{h \times w}$. Here, $m=1$ signifies the region to be inpainted/edited, and $m=0$ signifies the region to be preserved.
        \item Sample an initial full-noise latent variable: $z_T \sim \mathcal{N}(0, I)$, where $z_T \in \mathbb{R}^{h \times w \times c_z}$.
    \end{itemize}

    \item \textbf{Iterative Denoising (Reverse Diffusion)}: For $t = T, T-1, \dots, 1$:
    \begin{enumerate}
        \item \textbf{Prepare U-Net Input}: If the U-Net was trained for inpainting using channel concatenation (as in Eq.~\ref{eq:training_loss}), the input at step $t$ is:
        \begin{equation}
            z_t^{\text{U-Net\_input}} = \text{concat}(z_t, m, (1-m) \odot z_0^{\text{src}})
            \label{eq:unet_input_inference}
        \end{equation}
        The term $(1-m) \odot z_0^{\text{src}}$ provides the model with the clean latent information from the regions to be preserved.
        
        \item \textbf{Predict Noise (with Classifier-Free Guidance - CFG)}:
        The U-Net predicts the noise based on the current latent $z_t^{\text{U-Net\_input}}$, timestep $t$, and text conditioning $c_e$. Using CFG with a guidance scale $w_{\text{cfg}}$:
        \begin{align}
            \epsilon_{\text{cond}} &= \epsilon_\theta(z_t^{\text{U-Net\_input}}, t, c_e) \\
            \epsilon_{\text{uncond}} &= \epsilon_\theta(z_t^{\text{U-Net\_input}}, t, c_{\emptyset}) \quad \text{(where } c_{\emptyset} \text{ is a null-text/unconditional embedding)} \\
            \epsilon_{\text{pred},t} &= \epsilon_{\text{uncond}} + w_{\text{cfg}} \cdot (\epsilon_{\text{cond}} - \epsilon_{\text{uncond}})
            \label{eq:cfg_noise}
        \end{align}

        \item \textbf{Estimate Denoised Latent $\hat{z}_{0|t}$}: Based on $z_t$ and $\epsilon_{\text{pred},t}$, predict the "clean" latent:
        \begin{equation}
            \hat{z}_{0|t} = \frac{1}{\sqrt{\bar{\alpha}_t}} (z_t - \sqrt{1-\bar{\alpha}_t} \epsilon_{\text{pred},t})
            \label{eq:predict_z0}
        \end{equation}
        
        \item \textbf{Compute $z_{t-1}$ (DDIM-like step)}: The next latent state $z_{t-1}$ is computed. A common DDIM \cite{song2020denoising} step is:
        \begin{equation}
             z_{t-1}^{\text{model}} = \sqrt{\bar{\alpha}_{t-1}} \hat{z}_{0|t} + \sqrt{1 - \bar{\alpha}_{t-1} - \sigma_t^2} \cdot \epsilon_{\text{pred},t} + \sigma_t \tilde{\epsilon}_t
             \label{eq:ddim_step_model}
        \end{equation}
        where $\tilde{\epsilon}_t \sim \mathcal{N}(0, I)$ is fresh noise (for $\sigma_t > 0$) or $\tilde{\epsilon}_t = \epsilon_{\text{pred},t}$ if following certain interpretations. For deterministic DDIM, $\sigma_t = 0$. $z_{t-1}^{\text{model}}$ is the latent state fully generated by the model's prediction.

        \item \textbf{Enforce Preservation of Unmasked Regions (Blending/Resampling)}:
        To ensure that regions outside the edit mask $m$ remain unchanged from the source image $X_{\text{src}}$, the corresponding parts of $z_{t-1}^{\text{model}}$ are replaced with a noised version of $z_0^{\text{src}}$.
        First, compute the correctly noised version of the original unmasked regions for step $t-1$:
        \begin{equation}
            z_{t-1}^{\text{src\_noised}} = \sqrt{\bar{\alpha}_{t-1}} z_0^{\text{src}} + \sqrt{1 - \bar{\alpha}_{t-1}} \epsilon'_t
            \label{eq:src_noised}
        \end{equation}
        where $\epsilon'_t \sim \mathcal{N}(0, I)$ is new noise sampled at each step, or derived consistently if $\sigma_t=0$.
        Then, combine the model's prediction for the masked region with the original content for the unmasked region:
        \begin{equation}
            z_{t-1} = m \odot z_{t-1}^{\text{model}} + (1-m) \odot z_{t-1}^{\text{src\_noised}}
            \label{eq:blending_step}
        \end{equation}
        This step is crucial for inpainting, ensuring that only the specified regions $m$ are updated by the model's prediction, while other regions $(1-m)$ revert to the (appropriately noised) original source image latent.
    \end{enumerate}

    \item \textbf{Final Decoding}:
    After $T$ denoising steps, the resulting latent representation $z_0$ (which is $z_{t-1}$ from the last step where $t=1$) is decoded back into pixel space to obtain the final edited image $X_{\text{res}}$:
    \begin{equation}
        X_{\text{res}} = \mathcal{D}(z_0)
        \label{eq:final_decode}
    \end{equation}
\end{enumerate}
This inference process allows Stable Diffusion to perform targeted inpainting within the region $M_{\text{region}}$ guided by $t_e$, while preserving the content of $X_{\text{src}}$ outside this region. The blending step (Eq.~\ref{eq:blending_step}) is the primary mechanism for conditioning on the mask $M_{\text{region}}$ and the source image $X_{\text{src}}$ during the iterative generation.

\section{Additional Experiment}

\subsection{Image Editing Result Examples}
\begin{figure}[htbp]
    \centering

    % First column (3 images arranged horizontally)
    \begin{subfigure}[b]{0.49\textwidth}
        \centering
        \begin{minipage}[t]{0.32\textwidth}
            \centering
            \includegraphics[width=\linewidth]{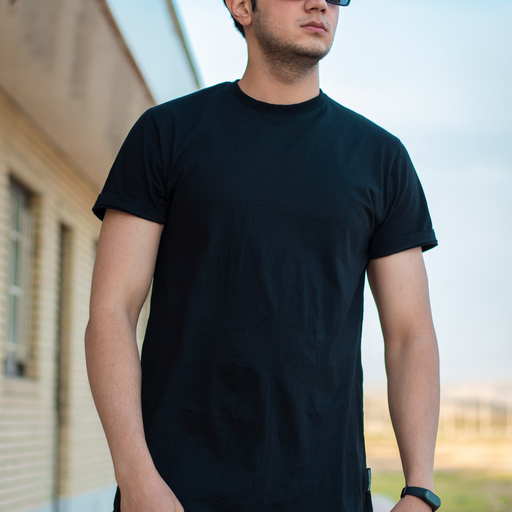}
            \caption*{\scriptsize A person wearing a t-shirt}
        \end{minipage}
        \hfill
        \begin{minipage}[t]{0.32\textwidth}
            \centering
            \includegraphics[width=\linewidth]{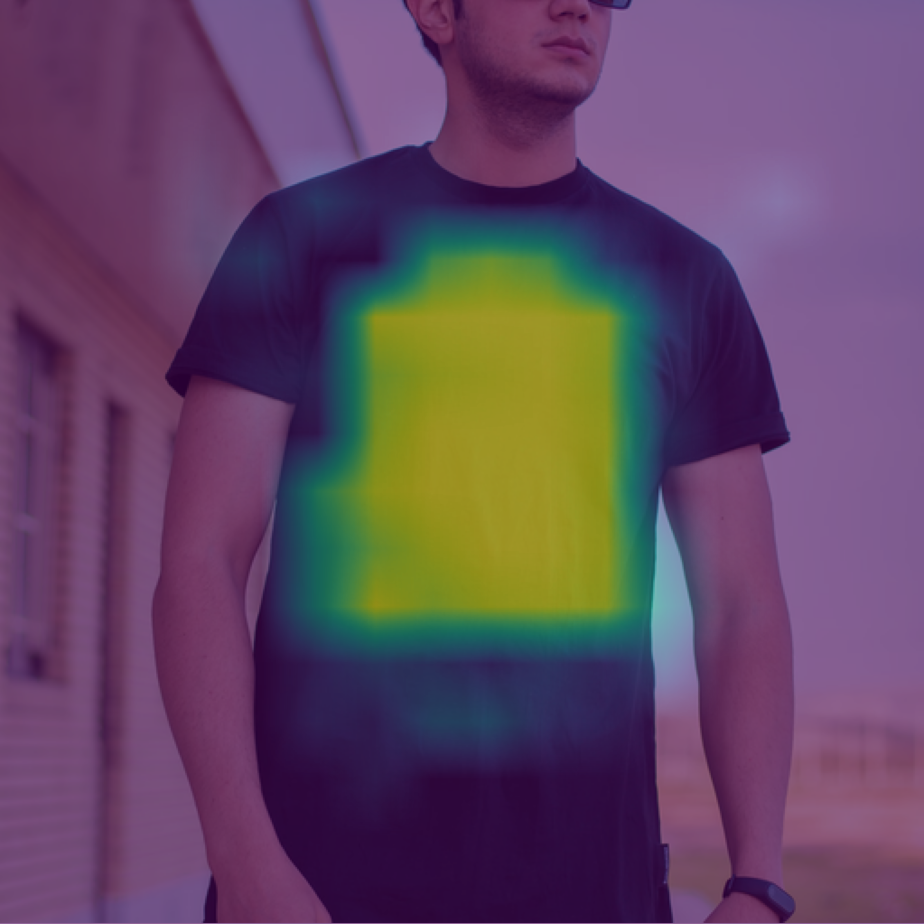}
            \caption*{\scriptsize Add a NASA logo on the t-shirt}
        \end{minipage}
        \hfill
        \begin{minipage}[t]{0.32\textwidth}
            \centering
            \includegraphics[width=\linewidth]{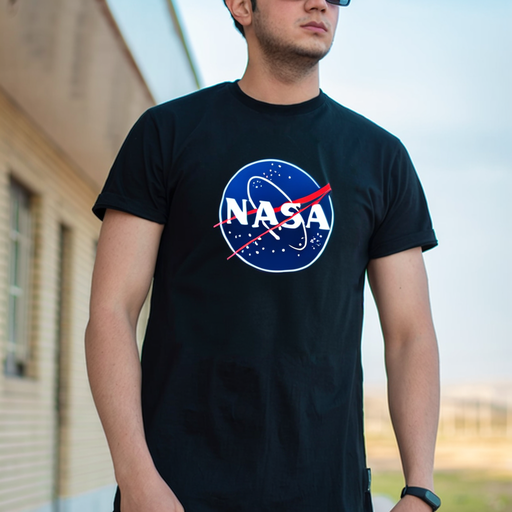}
            \caption*{\scriptsize A person wearing a t-shirt with the NASA logo}
        \end{minipage}
    \end{subfigure}
    \hfill
    % Second column (3 images arranged horizontally)
    \begin{subfigure}[b]{0.49\textwidth}
        \centering
        \begin{minipage}[t]{0.32\textwidth}
            \centering
            \includegraphics[width=\linewidth]{fig/13.png}
            \caption*{\scriptsize A tree and several rocks on the beach}
        \end{minipage}
        \hfill
        \begin{minipage}[t]{0.32\textwidth}
            \centering
            \includegraphics[width=\linewidth]{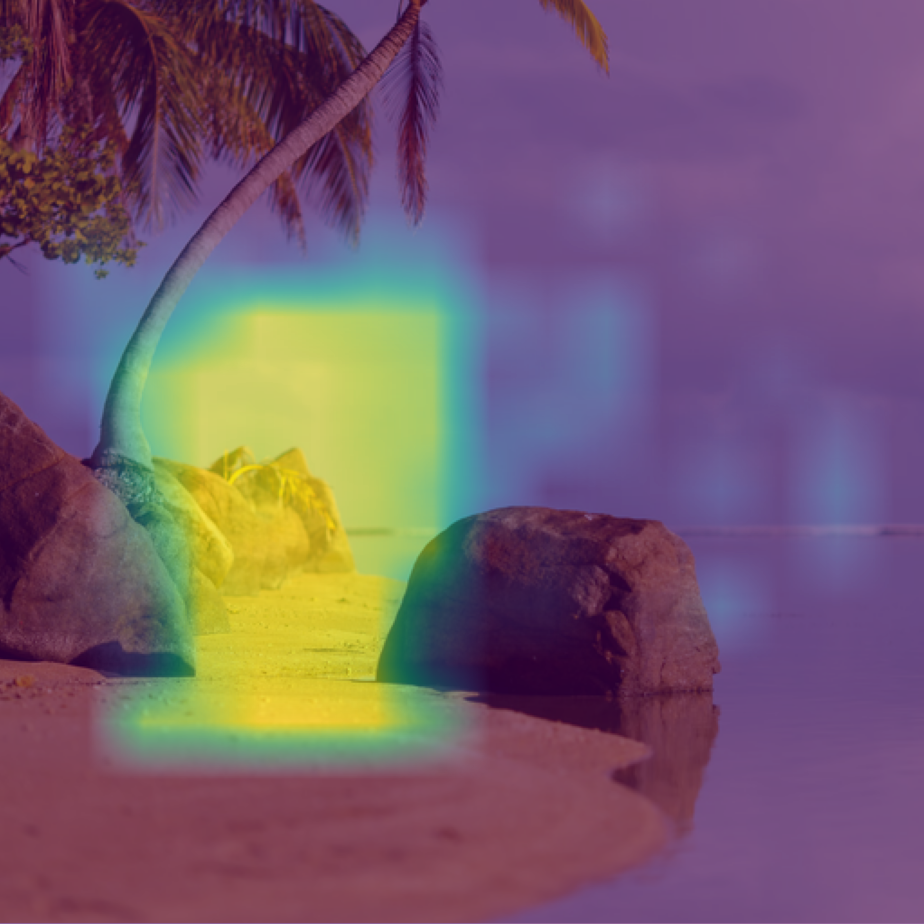}
            \caption*{\scriptsize Add a pedestrian walking along the beach}
        \end{minipage}
        \hfill
        \begin{minipage}[t]{0.32\textwidth}
            \centering
            \includegraphics[width=\linewidth]{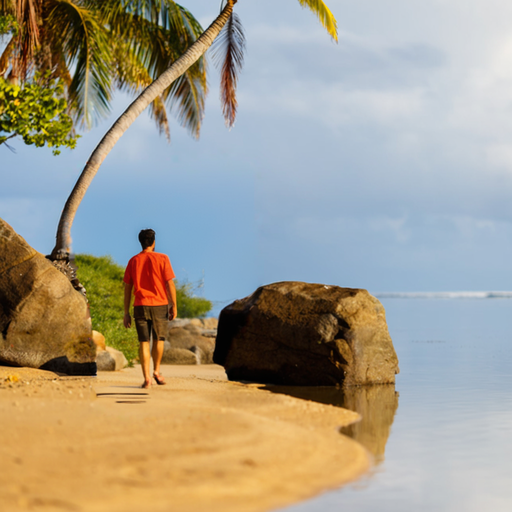}
            \caption*{\scriptsize A pedestrian walks along the beach near a tree and several rocks}
        \end{minipage}
    \end{subfigure}

    % First column (3 images arranged horizontally)
    \begin{subfigure}[b]{0.49\textwidth}
        \centering
        \begin{minipage}[t]{0.32\textwidth}
            \centering
            \includegraphics[width=\linewidth]{fig/10.png}
            \caption*{\scriptsize A plate of spaghetti with a fork}
        \end{minipage}
        \hfill
        \begin{minipage}[t]{0.32\textwidth}
            \centering
            \includegraphics[width=\linewidth]{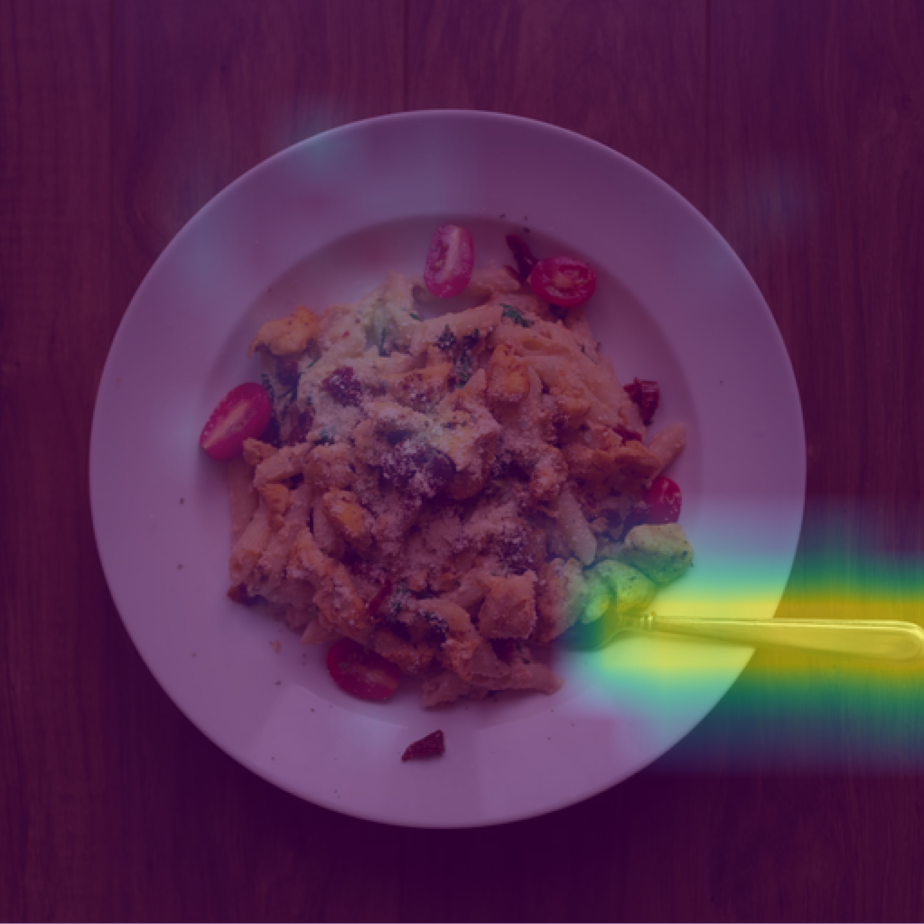}
            \caption*{\scriptsize Remove the fork}
        \end{minipage}
        \hfill
        \begin{minipage}[t]{0.32\textwidth}
            \centering
            \includegraphics[width=\linewidth]{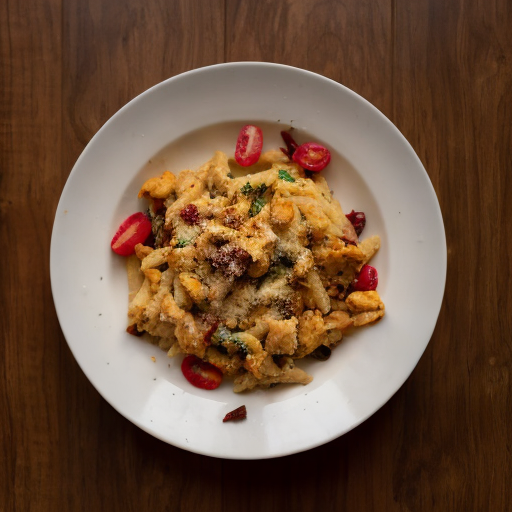}
            \caption*{\scriptsize A plate of spaghetti}
        \end{minipage}
    \end{subfigure}
    \hfill
    % Second column (3 images arranged horizontally)
    \begin{subfigure}[b]{0.49\textwidth}
        \centering
        \begin{minipage}[t]{0.32\textwidth}
            \centering
            \includegraphics[width=\linewidth]{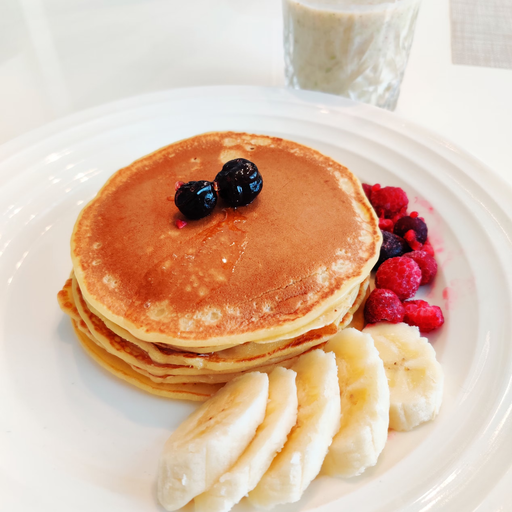}
            \caption*{\scriptsize Some pancakes with banana slices and berries on a plate}
        \end{minipage}
        \hfill
        \begin{minipage}[t]{0.32\textwidth}
            \centering
            \includegraphics[width=\linewidth]{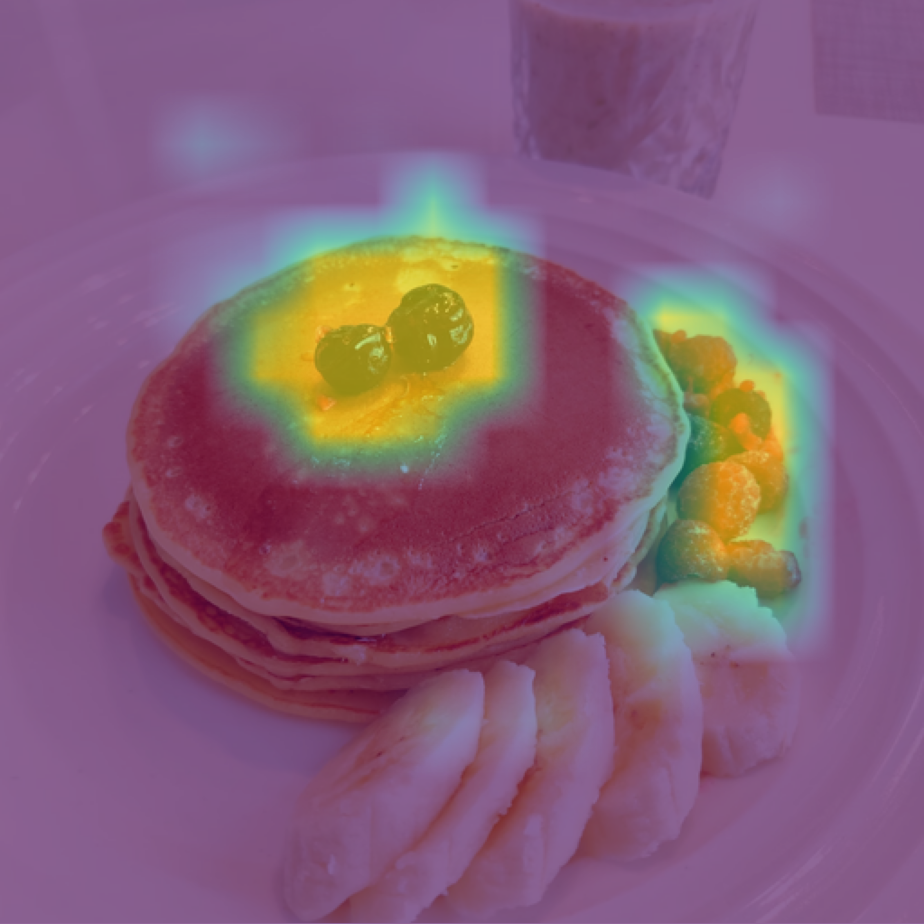}
            \caption*{\scriptsize Remove the berries}
        \end{minipage}
        \hfill
        \begin{minipage}[t]{0.32\textwidth}
            \centering
            \includegraphics[width=\linewidth]{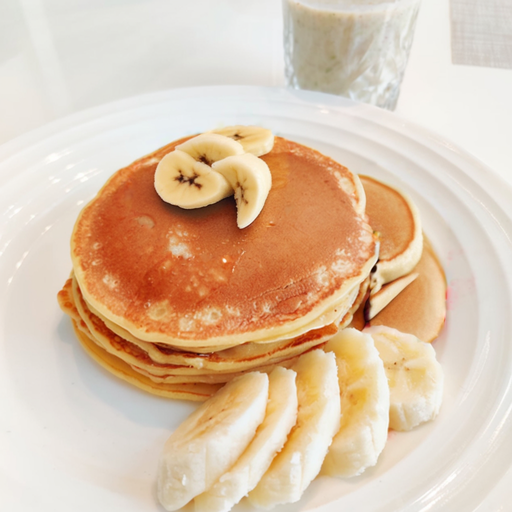}
            \caption*{\scriptsize Some pancakes with banana slices on a plate}
        \end{minipage}
    \end{subfigure}

    % First column (3 images arranged horizontally)
    \begin{subfigure}[b]{0.49\textwidth}
        \centering
        \begin{minipage}[t]{0.32\textwidth}
            \centering
            \includegraphics[width=\linewidth]{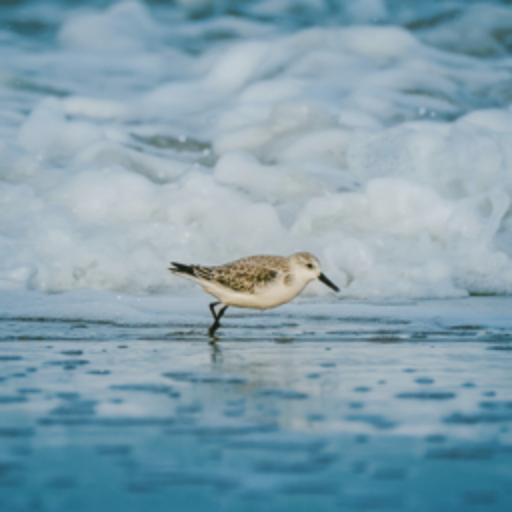}
            \caption*{\scriptsize A bird on a blue beach}
        \end{minipage}
        \hfill
        \begin{minipage}[t]{0.32\textwidth}
            \centering
            \includegraphics[width=\linewidth]{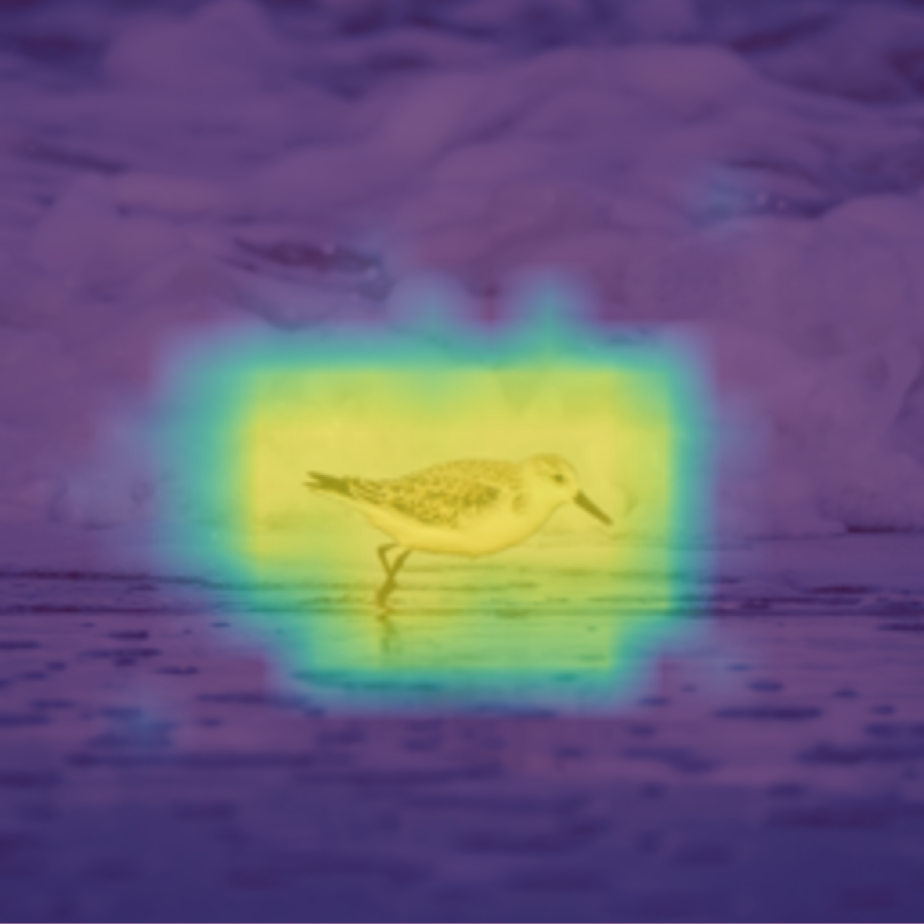}
            \caption*{\scriptsize Change the bird to a rabbit with white fur}
        \end{minipage}
        \hfill
        \begin{minipage}[t]{0.32\textwidth}
            \centering
            \includegraphics[width=\linewidth]{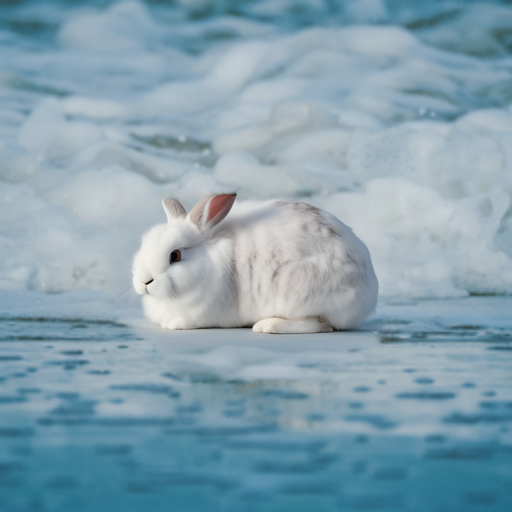}
            \caption*{\scriptsize a rabbit with white fur on a blue beach}
        \end{minipage}
    \end{subfigure}
    \hfill
    % Second column (3 images arranged horizontally)
    \begin{subfigure}[b]{0.49\textwidth}
        \centering
        \begin{minipage}[t]{0.32\textwidth}
            \centering
            \includegraphics[width=\linewidth]{fig/9.png}
            \caption*{\scriptsize An empty canvas}
        \end{minipage}
        \hfill
        \begin{minipage}[t]{0.32\textwidth}
            \centering
            \includegraphics[width=\linewidth]{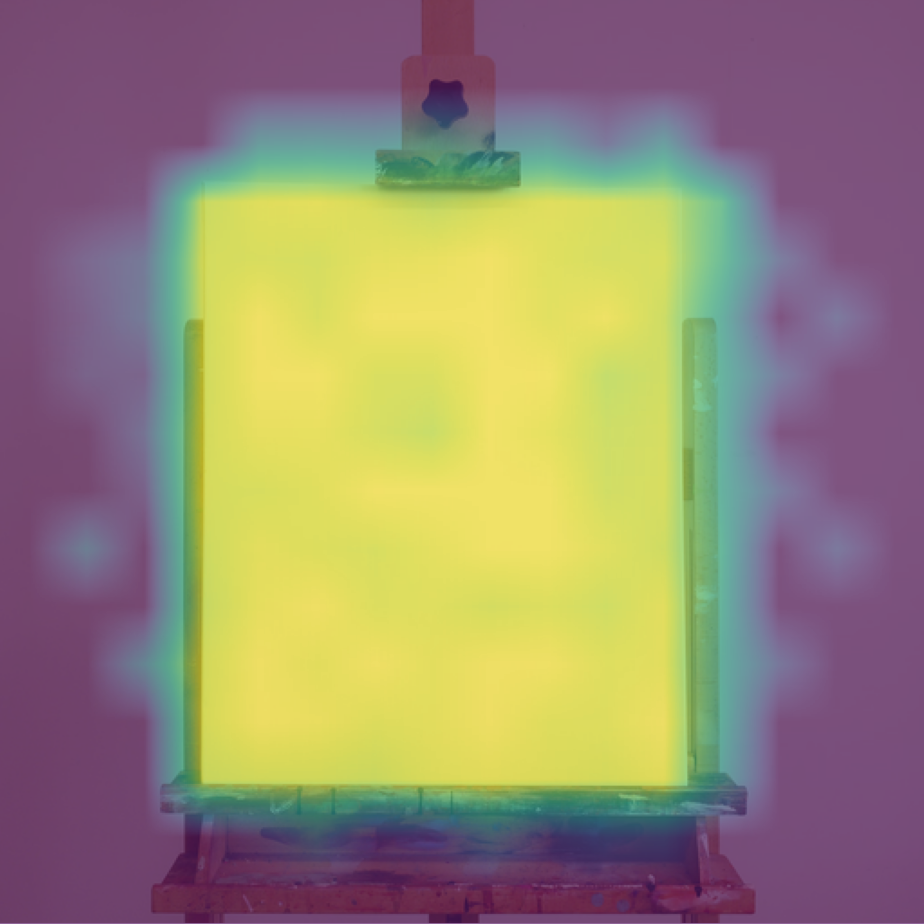}
            \caption*{\scriptsize Draw Leonardo da Vinci's 'Mona Lisa' to the canvas}
        \end{minipage}
        \hfill
        \begin{minipage}[t]{0.32\textwidth}
            \centering
            \includegraphics[width=\linewidth]{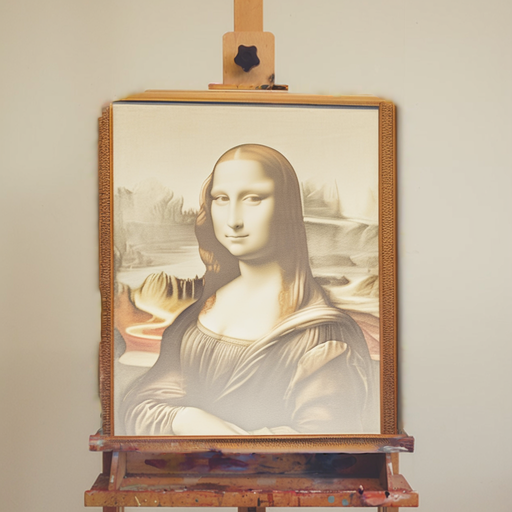}
            \caption*{\scriptsize Leonardo da Vinci's painting 'Mona Lisa'}
        \end{minipage}
    \end{subfigure}

    % First column (3 images arranged horizontally)
    \begin{subfigure}[b]{0.49\textwidth}
        \centering
        \begin{minipage}[t]{0.32\textwidth}
            \centering
            \includegraphics[width=\linewidth]{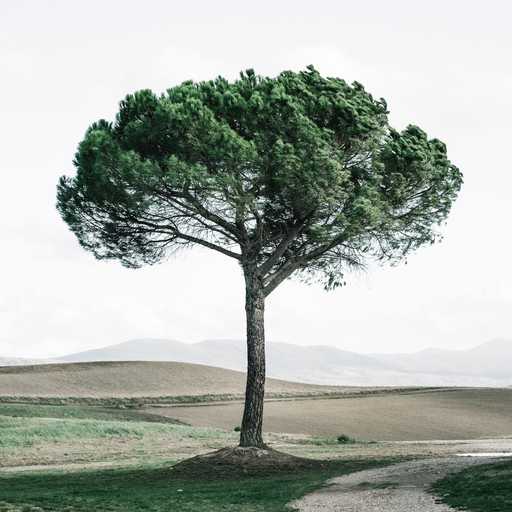}
            \caption*{\scriptsize A green tree}
        \end{minipage}
        \hfill
        \begin{minipage}[t]{0.32\textwidth}
            \centering
            \includegraphics[width=\linewidth]{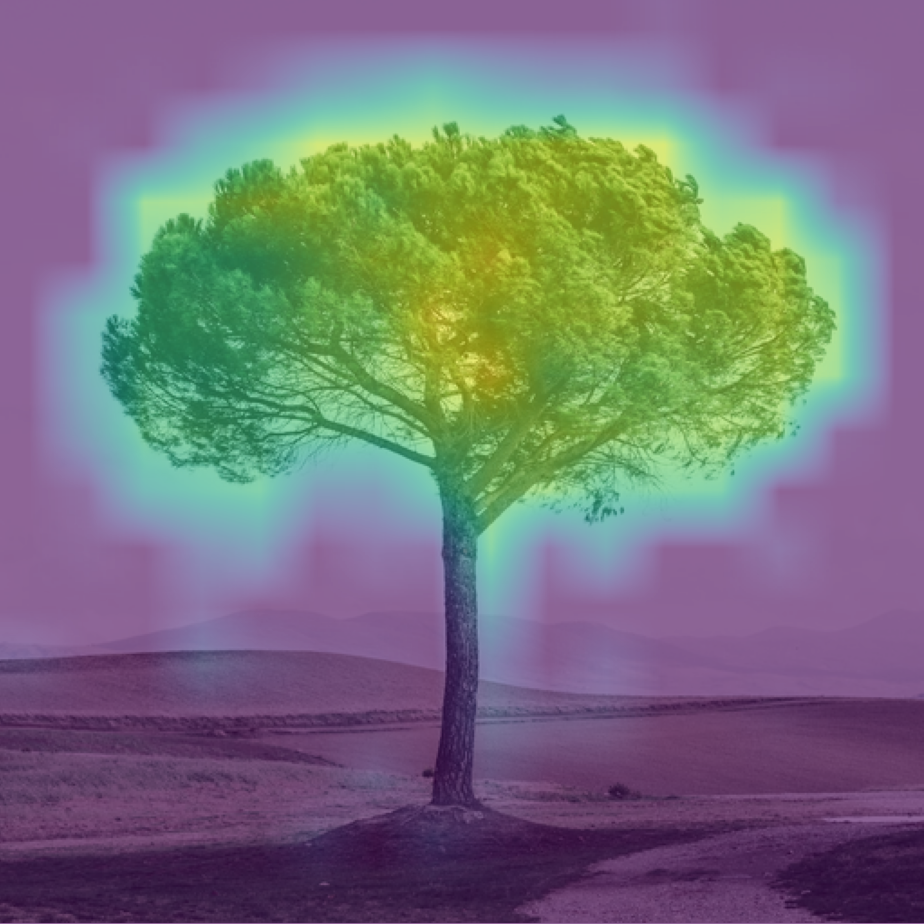}
            \caption*{\scriptsize Make the tree's pink cherry blossoms bloom}
        \end{minipage}
        \hfill
        \begin{minipage}[t]{0.32\textwidth}
            \centering
            \includegraphics[width=\linewidth]{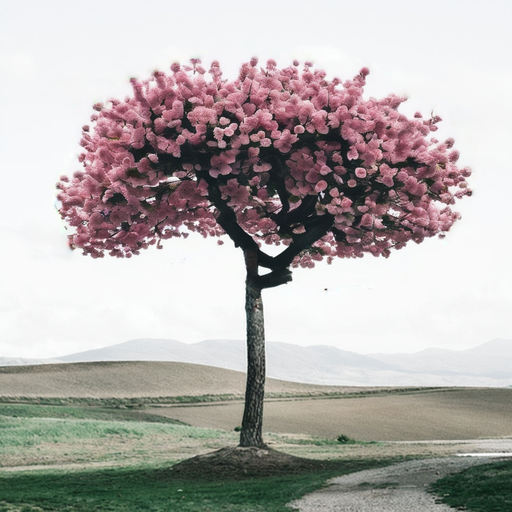}
            \caption*{\scriptsize A blooming pink cherry blossom tree}
        \end{minipage}
    \end{subfigure}
    \hfill
    % Second column (3 images arranged horizontally)
    \begin{subfigure}[b]{0.49\textwidth}
        \centering
        \begin{minipage}[t]{0.32\textwidth}
            \centering
            \includegraphics[width=\linewidth]{fig/17.png}
            \caption*{\scriptsize A cake topped with cookies}
        \end{minipage}
        \hfill
        \begin{minipage}[t]{0.32\textwidth}
            \centering
            \includegraphics[width=\linewidth]{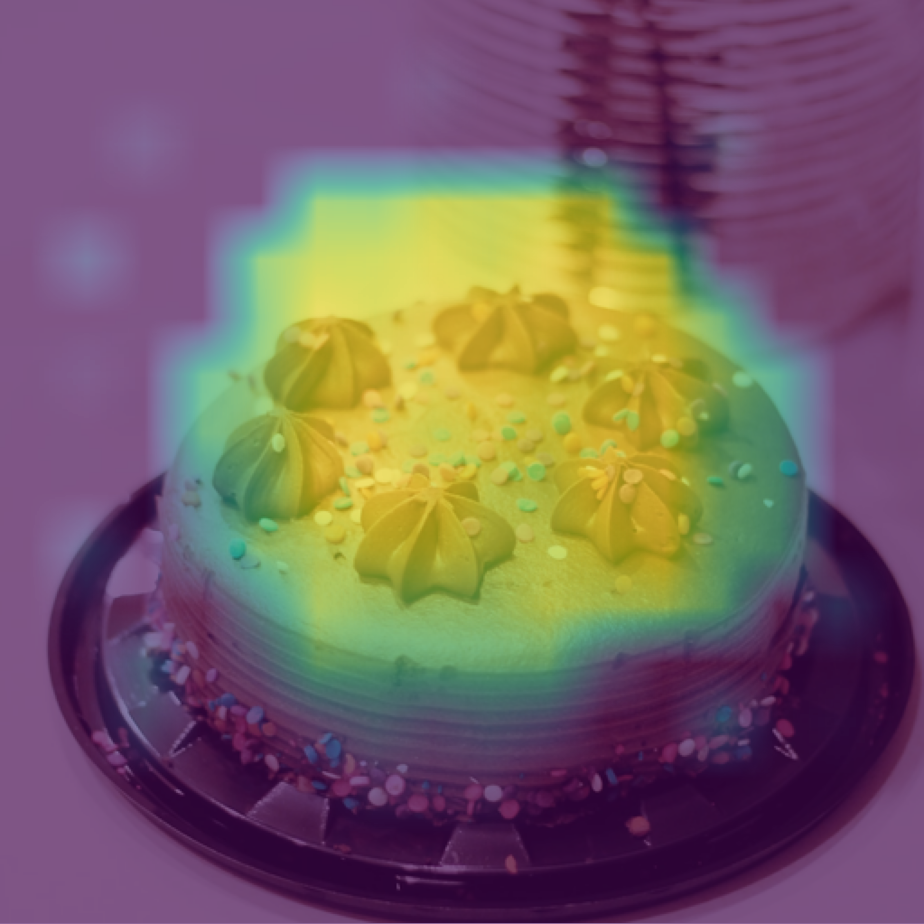}
            \caption*{\scriptsize Replace the cookies with candles}
        \end{minipage}
        \hfill
        \begin{minipage}[t]{0.32\textwidth}
            \centering
            \includegraphics[width=\linewidth]{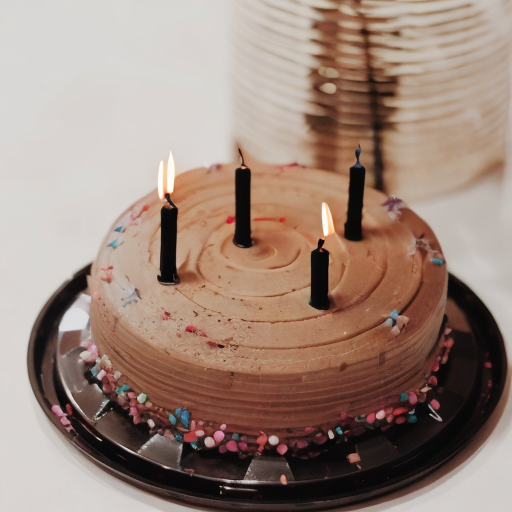}
            \caption*{\scriptsize A cake topped with candles}
        \end{minipage}
    \end{subfigure}

    \caption{\textbf{Instruction-driven image editing results.} The first row shows additive operations, the second row shows removal operations, and the third and fourth rows illustrate modification operations. Our method is capable of performing various image editing tasks while constraining changes within the learnable region.}
 % Main caption
    \label{fig:variousOperation}
    \vspace{-1em}
\end{figure}

We present more editing result examples.
The results shown in Figure~\ref{fig:variousOperation} demonstrate that our method can effectively handle a wide range of editing tasks across different contexts, producing high-quality edited images.

\subsection{Quantitative Ablation Study}
Beyond the qualitative ablation results shown in Figure~\ref{fig:ablation}, we also present quantitative results in Table~\ref{tab:ablation}, which demonstrate that all loss components are necessary for our method.

\begin{table}[ht]
\vspace{-1em}
\centering
\caption{\textbf{Ablation Study Results on the Emu Edit Test}. We report benchmark results for models trained with different loss configurations.}

\begin{tabular}{lccccc}
\toprule
\textbf{Setting} & \textbf{CLIPdir$\uparrow$} & \textbf{CLIPout$\uparrow$} & \textbf{L1$\downarrow$} & \textbf{CLIPimg$\uparrow$} & \textbf{DINO$\uparrow$} \\
\midrule
w/o $L_\text{semAlign}$~\eqref{eq:loss_semAlign} & 0.0706 & 0.2567 & 0.0789 & 0.8704 & 0.7921 \\
w/o $L_\text{CLIPg}$~\eqref{eq:loss_CLIPg} & 0.0768 & 0.2663 & 0.0824 & 0.8652 & 0.7712 \\
w/o $L_\text{CLIPd}$~\eqref{eq:loss_CLIPd} & 0.0772 & 0.2654 & 0.0843 & 0.8682 & 0.7724 \\
w/o $L_\text{CLIPs}$~\eqref{eq:loss_CLIPs} & 0.0878 & 0.2789 & 0.0913 & 0.8483 & 0.7437 \\
\rowcolor{gray!20}
\textbf{Ours}  & \textbf{0.1088} & \textbf{0.2842} & \textbf{0.0723} & \textbf{0.8913} & \textbf{0.8337} \\
\bottomrule
\end{tabular}
\label{tab:ablation}
\vspace{-1em}

\end{table}

\section{Experiments compute resources}
Our model is trained on a platform with 8 NVIDIA A100 80GB GPUs, using a batch size of 256. Each epoch takes approximately 4.6 hours under this setup.

\section{Limitation}
Due to inherent limitations of the CLIP model, certain complex editing instructions, such as ``Move the cat to the sofa," which require deeper image understanding, may fail. In dense scenes, our method may produce inaccurate editing regions, leading to suboptimal performance. Additionally, it inherits fairness issues from the underlying generative models.

%%%%%%%%%%%%%%%%%%%%%%%%%%%%%%%%%%%%%%%%%%%%%%%%%%%%%%%%%%%%

\end{document}